\documentclass[letterpaper,twocolumn,10pt]{article}
\usepackage{usenix2019_v3}

\usepackage{tikz}
\usepackage{amsmath}
\usepackage{algorithm}
\usepackage{algpseudocode}
\usepackage{setspace}
\usepackage{booktabs}
\usepackage{multirow}
\usepackage{adjustbox}
\usepackage{makecell}
\usepackage{amssymb}
\usepackage{hyperref}

\makeatletter
\renewcommand*{\@fnsymbol}[1]{\ensuremath{\ifcase#1\or *\or *\or
   \mathsection\or \mathparagraph\or \|\or **\or \dagger\dagger
   \or \ddagger\ddagger \else\@ctrerr\fi}}
\makeatother

\begin{document}
\title{\Large \bf \textit{Mirage in the Eyes}: Hallucination Attack on \\ Multi-modal Large Language Models with \textit{Only} Attention Sink}

\author{
{\rm Yining Wang$^{1}$, 
Mi Zhang$^{1,}$\thanks{ Corresponding authors.} , 
Junjie Sun$^{1}$, 
Chenyue Wang$^{1}$,
Min Yang$^{1,}$$^{\textcolor{green!80!black}{*}}$,}\\
{\rm Hui Xue$^{2}$, 
Jialing Tao$^{2}$,
Ranjie Duan$^{2}$,
Jiexi Liu$^{2}$
}\\
$^{1}$ Fudan University, $^{2}$ Alibaba Group\\
{\normalsize \{ynwang22@m., mi\_zhang@, jjsun22@m., wangcy23@m., m\_yang@\}fudan.edu.cn} \\
{\normalsize \{hui.xueh, jialing.tjl, ranjie.drj, liujiexi.ljx\}@alibaba-inc.com}
} 

\maketitle

\begin{abstract}
Fusing visual understanding into language generation, Multi-modal Large Language Models (MLLMs) are revolutionizing visual-language applications. Yet, these models are often plagued by the \textit{hallucination problem}, which involves generating inaccurate objects, attributes, and relationships that do not match the visual content. In this work, we delve into the internal attention mechanisms of MLLMs to reveal the underlying causes of hallucination, exposing the inherent vulnerabilities in the instruction-tuning process. 

We propose a novel \textit{hallucination attack} against MLLMs that exploits \textit{attention sink} behaviors to trigger hallucinated content with minimal image-text relevance, posing a significant threat to critical downstream applications. Distinguished from previous adversarial methods that rely on fixed patterns, our approach generates dynamic, effective, and highly transferable visual adversarial inputs, without sacrificing the quality of model responses. Comprehensive experiments on 6 prominent MLLMs demonstrate the efficacy of our attack in compromising black-box MLLMs even with extensive mitigating mechanisms, as well as the promising results against cutting-edge commercial APIs, such as GPT-4o and Gemini 1.5. Our code is available at \url{https://huggingface.co/RachelHGF/Mirage-in-the-Eyes}.

\end{abstract}
\section{Introduction}
Integrating visual comprehension into language models, Multi-modal Large Language Models \cite{liu2024visual, liu2024improved, Dai2023InstructBLIPTG, Chen2023ShikraUM, bai2023qwen} enable interaction with users across various modalities, and provide responses that demonstrate a deep understanding of complex visual semantics. Through instruction-tuning in multi-modal spaces, MLLMs have significantly advanced vision-language tasks like image captioning \cite{chen2023sharegpt4v, chen2024lion}, visual grounding \cite{peng2024grounding, peng2023kosmos}, and multi-modal conversations \cite{zhang2023llavar, liu2024mminstruct}.

Despite their revolutionary impact, MLLMs face a significant challenge: \textit{the hallucination problem}. This occurs when they generate irrelevant or entirely fabricated responses according to the image content. Examples include mentioning non-existent objects \cite{chen2024multi, li2023evaluating}, providing inaccurate attributes \cite{liu2024phd}, or describing inconsistent relationships between objects \cite{wang2023llm}. As MLLMs are increasingly employed to aid decision-making, task planning, and user interaction in critical fields like medical reasoning \cite{pal2024gemini, li2024llava}, autonomous driving \cite{ding2023hilm, xu2024drivegpt4}, and robotic manipulation \cite{li2024manipllm, long2024robollm}, their tendency to generate hallucinated responses poses significant risks. In applications with user-defined inputs, such as AI assistants \cite{nguyen2024yo}, customer services \cite{chen2024ipl}, and physical therapy tools \cite{zhou2024pre}, manipulated contents may be injected or spread online, compromising system reliability and potentially misleading users without domain expertise.

To address the unintended effects of unfaithful MLLMs, the causes of hallucinations have been explored \cite{bai2024hallucination}. Some suggest that hallucinations arise from the imbalance between weak vision models and powerful LLM backbones \cite{leng2024mitigating, guan2024hallusionbench, lee2023volcano}, which causes MLLMs to over-rely on language priors (e.g., the tendency of associating \textit{bananas} with the color \textit{yellow} more often than \textit{green}). Other studies point to the statistical biases in MLLM pre-training datasets \cite{leng2024mitigating, zhouanalyzing}, which often feature imbalanced object distributions and co-occurrence patterns, resulting in MLLMs generating irrelevant descriptions. While various studies have explored the causes of hallucination, they often focus on individual factors in isolation, largely in a post-hoc manner. In light of these gaps, our work explores the generation mechanisms of MLLMs, investigating the complex interactions between modalities to provide a comprehensive understanding of this phenomenon.

\begin{figure}[ht]
    \centering
    \vspace{-0.5em}
\includegraphics[width=0.98\columnwidth]{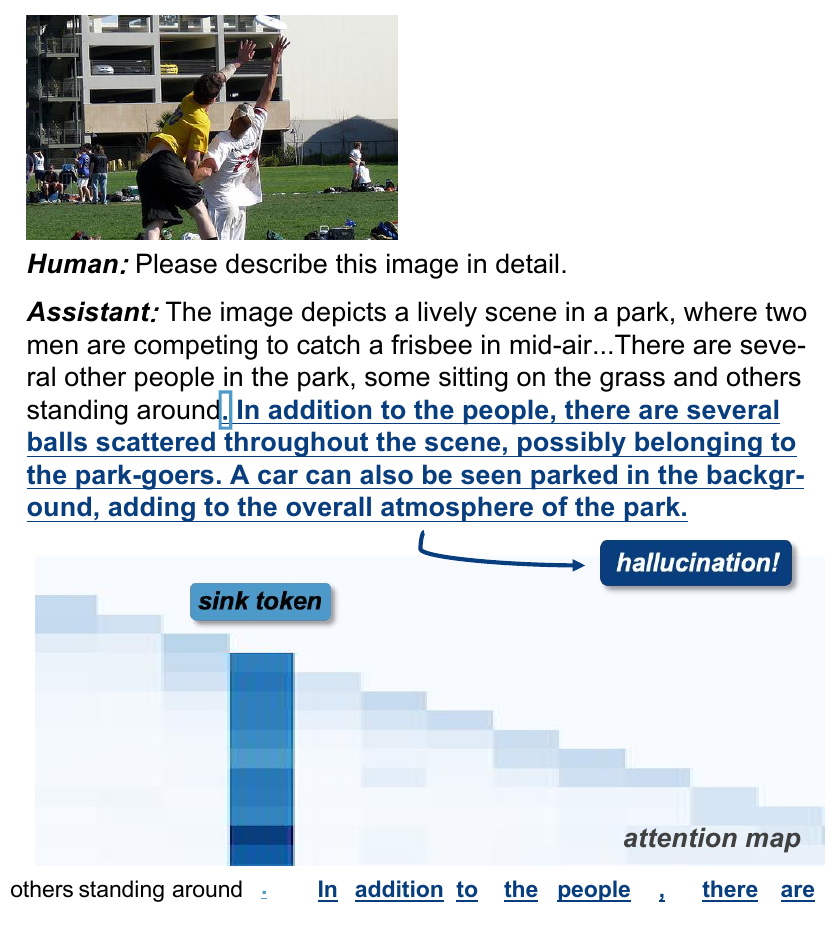}
\vspace{-1em}
\caption{An illustration of the attention sink phenomenon in MLLM responses. The sink token receives high attention scores in a columnar pattern. The hallucinated responses are marked bold with \textcolor[RGB]{8,61,126}{\textbf{indigo tokens}}. }
\label{fig:example}
\end{figure}

Recent research has identified a phenomenon known as \textit{attention sink} \cite{xiaoefficient, yu2024unveiling}, where certain tokens receive extremely high attention scores during the generation of LLM responses. Further research on MLLMs extends this concept to multi-modal settings \cite{huang2024opera}, showing that the sink token exhibits a unique columnar pattern within attention maps (as illustrated in Fig. \ref{fig:example}), drawing significant attention in the subsequent generative processes. Notably, the hallucinated outputs are observed to generally follow these sink tokens, indicating a potential connection between them. Through an in-depth examination of the instruction-tuning process in multi-modal training, we identify the critical flaw that MLLMs tend to produce irrelevant image-text content after following user instructions, while the aggregation of misleading global information further exacerbates the divergence from the actual image content.

Uncovering the formation of attention sink in hallucinated responses, we propose the first-ever \textit{hallucination attack} against MLLMs with \textit{only} attention sink. This pioneering attack is designed to exacerbate hallucinations in MLLM responses while maintaining their overall quality and utility. Our proposed attack overcomes the constraints of existing adversarial attacks against LLMs and MLLMs, which depend heavily on predefined target responses and task-specific datasets. By manipulating attention scores and hidden embeddings to induce sink tokens, our attack constructs dynamic, highly effective, and black-box transferable adversarial visual inputs. This method not only circumvents current mitigation strategies for hallucinations but also shows significant impacts on the latest commercial MLLM APIs. We hope this hallucination attack will expose the critical vulnerability of MLLMs in downstream applications, and drive advancements toward more reliable and high-performing multi-modal models.

\noindent \textbf{Our Distinction from Previous Attacks. } 
Adversarial attacks on MLLMs have been a concern since their development. These attacks aim to provoke harmful model responses \cite{niu2024jailbreaking, ma2024visual, shayegani2023jailbreak, cheng2024typography} or to impair model performance on specific tasks \cite{gao2024adversarial, cui2024robustness}, but they still rely heavily on predefined patterns or task-specific datasets. For instance, perturbation-based attacks \cite{niu2024jailbreaking, zhao2024evaluating, qi2024visual} optimize adversarial perturbations according to predefined target responses, which require massive human labor in crafting and filtering. Other text-based attacks insert human-written jailbreak templates \cite{ma2024visual, luo2024jailbreakv} (e.g., role-playing scenarios) to disrupt the safety alignments, but often lack transferability across different models. Recent structure-based attacks \cite{shayegani2023jailbreak, cheng2024typography} embed harmful instructions into images with typography or text-to-image models, which also require carefully designed textual triggers to redirect the model's attention. Concentrating on adversarial visual inputs, our work is distinct from previous adversarial attacks in two key aspects:
\begin{enumerate}
    \item \textbf{Attack Objective}: Our hallucination attack aims to induce the generation of erroneous objects, attributes, and relationships in MLLM responses, whereas previous adversarial attacks primarily focus on triggering harmful outputs and bypassing safety alignments. 
    \item \textbf{Attack Efficiency}: We achieve a dynamic and effective attack by directly manipulating the self-attention mechanisms during the MLLM generation process.
This approach overcomes the limitations of previous methods, which demand substantial human efforts to define the target behaviors of models.
\end{enumerate}

\noindent \textbf{Our Contributions}  are summarized as follows.
\begin{itemize}
    \item We present a comprehensive analysis linking the attention sink phenomenon with hallucination issues in MLLMs. By probing the inherent limitations of the instruction-tuning process, we expose the model’s tendency to produce two-segment responses with declined image-text relevance, where the aggregation of misleading information contributes to the hallucinated outputs.
    \item We propose the first hallucination attack targeting MLLMs with only the manipulation of attention sinks. The crafted adversarial visual inputs significantly exacerbate the object, attribute, and relationship hallucination without degrading response quality. By manipulating the attention mechanism and hidden embeddings, our attack achieves high transferability and adaptability without relying on predefined patterns.
    \item The extensive evaluation, assisted by GPT-4 \cite{achiam2023gpt}, assesses our attack in hallucination and response quality on six prominent MLLMs. Remarkable results demonstrate that our attack transfers effectively to black-box MLLMs and commercial APIs such as GPT-4o \cite{gpt4o} and Gemini 1.5 \cite{gemini}, successfully overcoming three categories of existing mitigation strategies. Up to 10.90\% and 12.74\% increase in hallucinated sentences and words highlight the vulnerability of critical downstream applications to our proposed attack.  
\end{itemize}

\section{Related Work}
\subsection{Multi-modal Large Language Models}
The surge of LLMs has highlighted their remarkable capabilities in in-context learning, instruction following, and multi-step reasoning \cite{minaee2024large}. MLLMs build on these strengths by incorporating additional modalities such as image, video, and audio, enabling users to engage with both textual and multi-modal prompts. MLLMs typically consist of three key components: multi-modal encoders (e.g., Vision Transformers \cite{dosovitskiy2020image} as visual encoders), pre-trained LLM backbones, and the adapter modules for modality alignment. The forerunner MLLMs like Flamingo \cite{alayrac2022flamingo} and MM-GPT \cite{gong2023multimodal} achieve cross-modality alignment by integrating gated cross-attention blocks within their LLM backbones, but are constrained by massive computational demands \cite{caffagni2024r}. On the other hand, models like LLaVA \cite{liu2024visual} and Shikra \cite{Chen2023ShikraUM} utilize linear projection layers to map multi-modal features into textual spaces, offering a more computationally efficient solution. Moreover, advanced MLLMs such as InstructBLIP \cite{Dai2023InstructBLIPTG}, mPLUG-owl \cite{ye2023mplug}, and Qwen-VL \cite{bai2023qwen} incorporate Q-former modules, which consist of two Transformer blocks with shared self-attention layers. These modules update a set of learnable queries to effectively integrate both textual and multi-modal features. Most MLLMs follow a two-stage training paradigm: first pre-training on large-scale datasets to bridge modality gaps, and then instruction-tuning on task-related data to enhance multimodal conversational capabilities.

\subsection{Mitigating Hallucination in MLLMs}
\label{sec:2.2}
The hallucination problem in MLLMs causes cross-modal inconsistencies, resulting in discrepancies between generated text responses and provided visual content \cite{bai2024hallucination}. Recent efforts to enhance their faithfulness fall into three categories: mitigation through \textit{decoding}, \textit{model retraining}, and \textit{post-processing}. 

Some work optimizes the decoding strategy during the inference stage, to suppress the generation of hallucinated responses \cite{huang2024opera, leng2024mitigating, han2024skip, chen2024alleviating}. For instance, OPERA \cite{huang2024opera} mitigates hallucination by penalizing columnar attention patterns during beam search decoding, thereby reducing the model's over-reliance on certain summary tokens. Meanwhile, VCD \cite{leng2024mitigating} calibrates model output distributions with both clean and distorted inputs, effectively counteracting the language priors of LLM backbones. 
In the realm of model retraining, HACL \cite{jiang2024hallucination} employs contrastive learning of multi-modal representations to distinguish between hallucinated and factual responses, while \cite{yue2024less} adjusts the prediction of EOS token to prevent excessively long hallucinated texts. Other retraining-based methods gather high-quality visual instruction tuning data to correct statistical biases\cite{yu2024hallucidoctor,  liu2023mitigating}, or resort to human feedback as a reward for response correction \cite{yu2024rlhf, sun2023aligning}.
Another category of work leverages post-hoc techniques to rectify hallucinations, such as training additional hallucination revisors \cite{zhouanalyzing, lee2023volcano, yin2023woodpecker}, or implementing multi-agent debate to examine the faithfulness of model responses \cite{lin2024interpreting}.

\subsection{Attention Sink}
Transformer-based models leverage the self-attention mechanism to capture global information \cite{vaswani2017attention}. However, during inference, it has been observed that some tokens receive disproportionately high attention compared to others, which is known as the \textit{attention sink} phenomenon. 

The phenomenon was first observed in Large Language Models (LLMs) \cite{xiaoefficient}, where initial tokens, despite their low semantic significance, receive high attention. One reasonable explanation focuses on the sum-to-one property of the Softmax operation within the self-attention mechanism, where redundant attention is allocated to the initial tokens, as they are always visible during the generation process. The phenomenon has spurred several innovative approaches in LLM quantization \cite{son2024prefixing} and LLM attention calibration \cite{yu2024unveiling}. 

When the attention mechanism is applied across different modalities, the attention sink is observed to cause hallucination responses \cite{huang2024opera}. This occurs when certain sink tokens, characterized by columnar patterns with high attention scores, dominate the prediction of subsequent tokens, overshadowing all preceding responses (as depicted in Fig. \ref{fig:example}). As a result, the models will neglect image tokens and prior context, ultimately resulting in hallucination responses.
\begin{figure*}[h]
    \centering
\includegraphics[width=\textwidth]{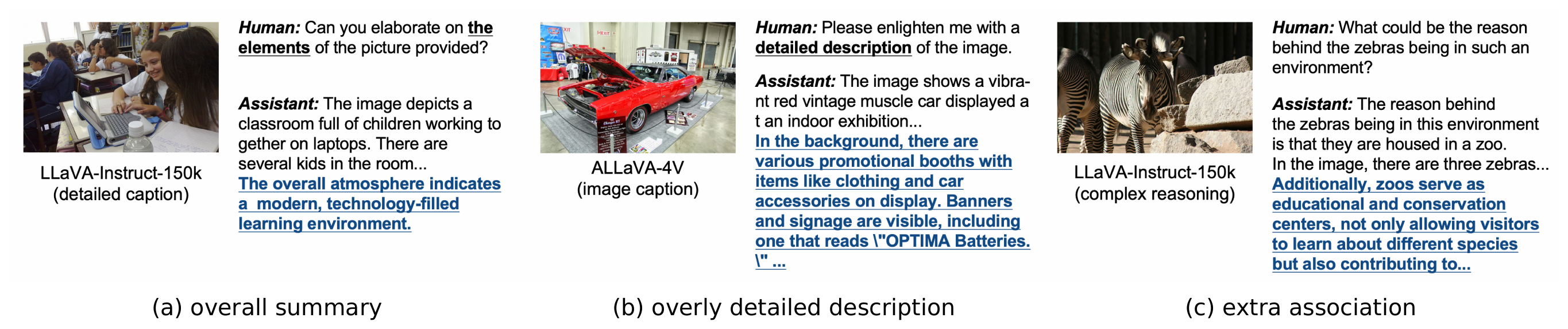}
\vspace{-1.8em}
\caption{Examples of inferred elements in ground truth responses: \textbf{(a)} overall summary of the image content, \textbf{(b)} overly detailed description of trivial objects, and \textbf{(c)} extra association not instructed by the task. The texts after "\textbf{\textit{Human:}}" denote instructions, and those after "\textbf{\textit{Assistant:}}" are ground truth responses. The examples are selected from the LLaVA-Instruct-150k \cite{liu2024visual} and ALLaVA-4V \cite{chen2024allava} datasets of detailed image caption and complex reasoning tasks, which are generated with GPT-4 \cite{achiam2023gpt} and GPT-4V \cite{gpt4v} models respectively. }
\label{fig:caption}
\vspace{-0.7em}
\end{figure*}

\section{Attention Sink and Hallucination}
\label{sec:3}
In the following section, we analyze the dynamics behind attention sink that relates to MLLM hallucinations. An in-depth investigation into the instruction-tuning stage of training reveals that, attention sink appears at the turning point of image-text relevance in model responses (Section \ref{sec:3.1}), which contains misleading global information that triggers subsequent hallucination content (Section \ref{sec:3.2}). 

\subsection{Analysis of Instruction-tuning Datasets}
\label{sec:3.1}
To enable user conversation with both text and image inputs, MLLMs are first pre-trained on large-scale datasets containing image-text pairs \cite{chen2023sharegpt4v, schuhmann2022laion}, and then fine-tuned on instruction datasets tailored for downstream applications. The instruction-tuning datasets consist of task descriptions and task-specific input-output pairs, covering multi-modal tasks such as image captioning \cite{chen2024allava}, visual question answering (VQA) \cite{liu2024visual, li2024monkey}, and referring expression comprehension (REC) \cite{chen2023minigpt}. For example, LLaVA \cite{liu2024visual} uses GPT-4 \cite{achiam2023gpt} to generate instruction-following dialogues, by providing it with captions and bounding boxes of COCO \cite{lin2014microsoft} images. The resulting dataset, LLaVA-Instruct-150k, has been utilized to fine-tune MLLMs like LLaVA \cite{liu2024visual}, Shikra \cite{Chen2023ShikraUM}, and InstructBLIP \cite{Dai2023InstructBLIPTG}.

Although instruction-tuning datasets include fine-grained question-answering pairs, the text-image relevance in model responses shows a decreasing trend. A closer examination of the ground truth responses reveals that, after describing the image content and following the instructions, the responses generally include additional inferred elements, such as overall summaries, overly detailed descriptions, and extra associations based on the image content, as displayed in Fig. \ref{fig:caption}. This may be attributed to the fact that models like GPT-4 \cite{achiam2023gpt} and GPT-4V \cite{gpt4v}, which are used for data generation, have strong comprehension and associative abilities. As a result, they tend to offer extra references and details in a user-friendly manner.

\begin{figure*}[h]
    \centering
\includegraphics[width=0.95\textwidth]{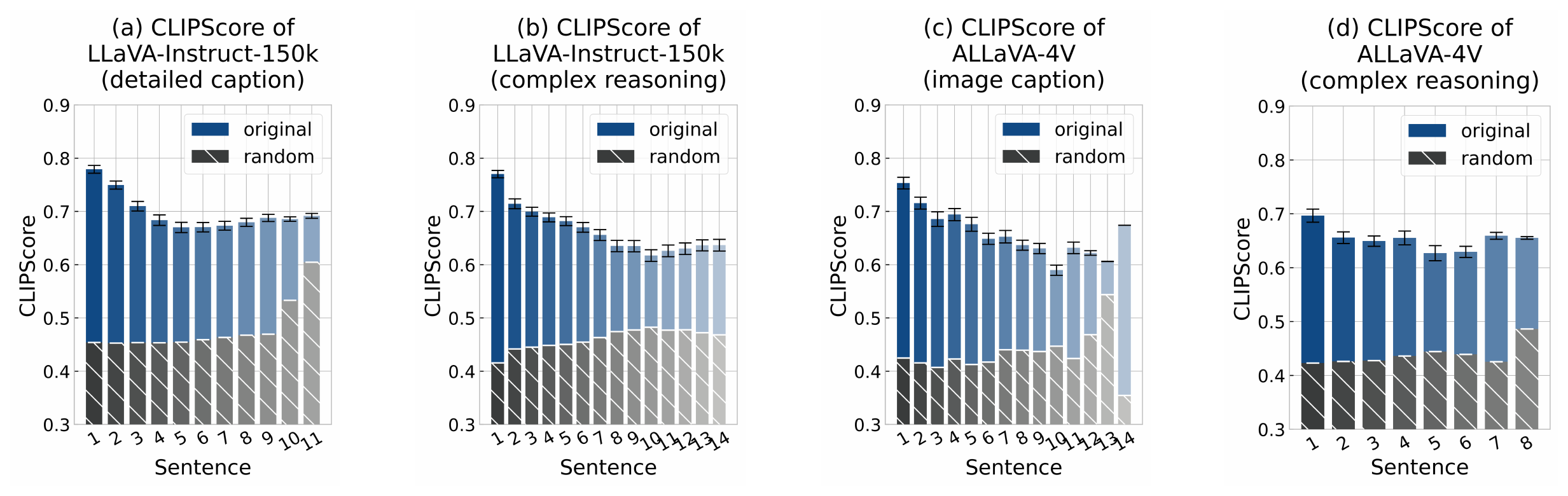}
\vspace{-1.5em}
\caption{Per-sentence CLIPScore between input images and ground truth responses in instruction-tuning datasets. 
We report CLIPScore between input images and random response sentences as the baseline, denoted as \textit{random}.}
\label{fig:clipscore}
\end{figure*}

To illustrate the decreasing text-image relevance in model responses of open-source instruction-tuning datasets, we select CLIPScore \cite{hessel2021clipscore} as a metric. The CLIPScore is generally adopted to evaluate the image–text compatibility \cite{liu2023models,tan2024empirical}, which first extracts the embeddings for both visual and textual inputs with CLIP \cite{radford2021learning} model, and then calculates the cosine similarity between these embeddings to reveal their relevance. We compute the CLIPScore between the input images and each sentence in the ground truth responses of the LLaVA-Instruct-150k \cite{liu2024visual} and ALLaVA-4V \cite{chen2024allava} datasets. Fig. \ref{fig:clipscore} reveals that, the ground truth responses exhibit a significant decrease in image-text relevance after the first few sentences. It results in two distinct segments in model-generated responses: (1) first the detailed descriptions closely tied to the image, and (2) content that is either loosely related to the image or beyond the visual interpretability of MLLMs.

\begin{figure*}[h]
    \centering
\includegraphics[width=\textwidth]{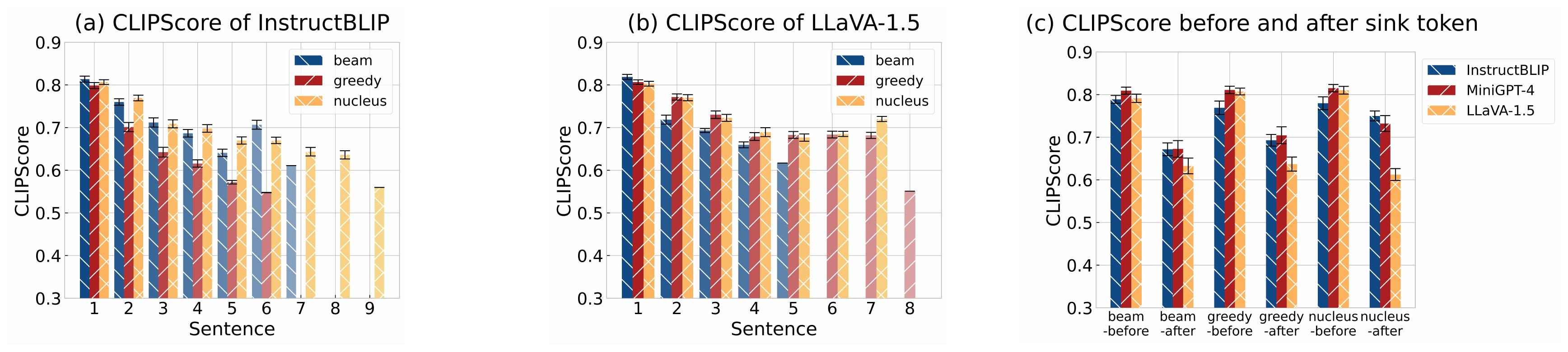}
\vspace{-1.5em}
\caption{\textbf{(a)}-\textbf{(b)} Per-sentence CLIPScore between input images and MLLM responses of InstructBLIP and LLaVA-1.5. \textbf{(c)} Mean CLIPScore of MLLM responses before and after the sink token. The postfix -\textit{beam}, -\textit{greedy}, and -\textit{nucleus} represent beam search, greedy search, and nucleus sampling decoding respectively. The missing bars indicate no generated sentences of the corresponding length.}
\label{fig:model}
\vspace{-0.5em}
\end{figure*}

The innate problem of datasets contributes to the hallucination problems of released MLLMs. When fine-tuned on such datasets, MLLMs tend to adopt the pattern of two-segment responses, first describing the image and then generating associative content. Moreover, when trained to fit the second part of the responses, MLLMs are compelled to generate details that they cannot visually comprehend \cite{yue2024less}, or abstract statements unrelated to the instructions. We also observe that the attention sink phenomenon emerges at the turning point of image-text relevance, which generally leads the hallucination responses with loose relation with images. We discover the following properties of attention sink originating from the instruction-tuning training:

\noindent \textbf{(1) MLLMs inherit the \textit{two-segment response} pattern from instruction-tuning datasets.} We prompt MLLMs to generate detailed image captions for VG 100K \cite{krishna2017visual} dataset, and evaluate the per-sentence CLIPScore between input images and their responses, as shown in Fig. \ref{fig:model} (a)-(b). Similar to the trend observed in instruction-tuning datasets, the MLLM responses clearly show a significant decline in image-text relevance, which applies to all three decoding strategies.

\noindent \textbf{(2) Attention sink appears at the turning point of CLIPScore.} By identifying the columnar patterns within the attention maps, we trace the presence of sink tokens and evaluate the mean CLIPScore of model responses before and after them, as shown in Fig. \ref{fig:model} (c). Our findings reveal that the attention sink appears to segment the response, with a marked decrease in image-text CLIPScore following the sink token, which suggests less relevant content and the prone to hallucinations. Notably, this issue is observed not only on models that are instruct-tuned on datasets displaying these tendencies (e.g., InstructBLIP and LLaVA-1.5 trained on LLaVA-Instruct-150k), but is also prevalent on MLLMs like MiniGPT-4, which are trained on closed-source datasets. This observation highlights a widespread problem across existing instruction-tuning paradigms.

\subsection{Aggregated information in Attention Sink}
\label{sec:3.2}
To explain the emergence of attention sink at the turning point of image-text relevance, we dig deeper into the attention mechanism during MLLM generation. We notice that, besides the high attention scores and columnar patterns, sink tokens are predominantly non-content tokens (e.g., punctuation marks and article words) that convey minimal semantic meaning. For instance, in the responses of LLaVA-1.5, up to 73.5\% of the sink tokens are non-content, indicating a tendency of allocating high attention to these semantically trivial elements.

We related this observation with a unique behavior discovered in Transformer-based models: \textit{the aggregation of knowledge}. The process occurs when global information of inputs is aggregated into uninformative tokens, providing a shortcut for the subsequent generation or classification. The phenomenon is observed in Transformer-based models like Vision Transformers (ViTs) \cite{dosovitskiy2020image}, LLMs, and MLLMs. For example, in language models, information is aggregated into functional label words (e.g., words like \textit{positive} and \textit{negative} in the task of sentiment analysis) in shallow layers to support final predictions \cite{wang2023label}. Similarly, in ViTs, where image patches are treated as tokens, the models inject global information into some background tokens to replace their local information, which facilitates the training of linear models for classification \cite{darcetvision}. In the study of MLLM hallucination, \cite{huang2024opera} also hypothesizes that certain tokens in MLLM responses aggregate crucial knowledge from contexts, and over-reliance on these tokens can lead to a neglect of the entire image content.

\begin{figure*}[ht]
    \centering
\includegraphics[width=0.9\textwidth]{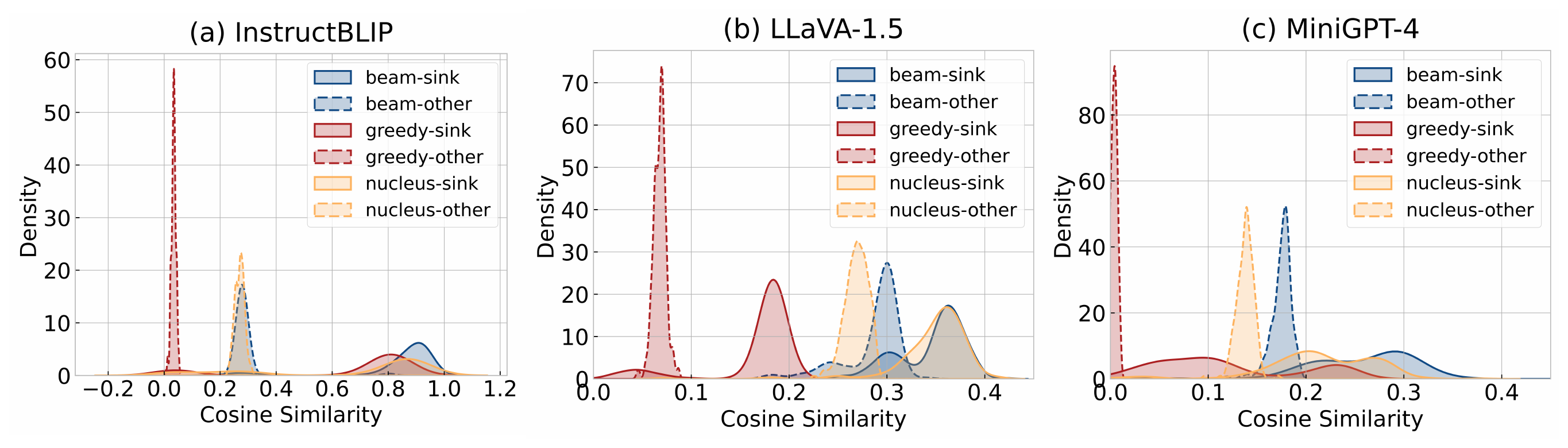}
\vspace{-1.5em}
\caption{Distribution of cosine similarity between multi-modal input embeddings and generated token embeddings. We compare the similarity of sink tokens (with the postfix \textit{-sink}) and all other tokens (with the postfix \textit{-other}).}
\label{fig:density}
\end{figure*}

Leading by the common phenomenon of aggregating behaviors, we note that part of the global information in MLLM, representing visual and textual inputs, is also aggregated into sink tokens. Fig. \ref{fig:density} presents a distribution of cosine similarity between the middle-layer embeddings of multi-modal inputs and the generated tokens. It's notable that sink tokens, which appear at the turning points of CLIPScore, exhibit a significantly higher resemblance to global input information compared to other tokens. We relate this observation to the hallucinated generation, and make the following analysis. 

\noindent \textbf{(1) Attention sinks aggregate information as global context.} The aggregating behavior of Transformer-based models is formed naturally during training, with sink tokens receiving high attention scores to aid in subsequent prediction or generation. In Fig. \ref{fig:density}, the higher similarity to input embeddings indicates that global multi-modal information is partly integrated into the sink tokens. In the generation process of MLLMs, multi-modal input tokens are positioned before the entire response, serving as a global context. We hypothesize that, inheriting the two-segment response pattern (Section \ref{sec:3.1}), attention sinks are chosen to distinguish between segments with different focus, content, and style in MLLM generation. This mechanism provides a more relevant global context for the latter part of the model's responses, minimizing the need for long-distance attention and aligning with the observed MLLM generation patterns.

\noindent \textbf{(2) Misleading aggregation triggers hallucinated response.} While the aggregation process aligns with the generation pattern of MLLMs, we note that only part of the global information is fused into sink tokens, which deviates from the original global information. We speculate that it is still due to deficiencies in the instruction tuning phase, where the second part of responses in training data often includes irrelevant descriptions (Section \ref{sec:3.1}), and will mislead the aggregating process with partial, trivial, and even wrong global information. Furthermore, the aggregation of global context into a single token inevitably results in a significant loss of information, diminishing the factual accuracy of the image content. Consequently, MLLMs are trained to aggregate misleading information as context for irrelevant generations. The high attention scores assigned to these sink tokens exacerbate the hallucination problem, introducing irrelevant objects, confused attributes, and incorrect relationships.

\section{Our Hallucination Attack}
\subsection{Motivation}
Based on the overall analysis in Section \ref{sec:3}, we highlight two important properties of attention sink in MLLM hallucination. First, the emergence of attention sink does not depend on specific textual or visual inputs, as the reason behind hallucination is deeply rooted in the attention mechanism during generation. Second, it requires no prior knowledge or external reference for detection, only the attention maps during the generation process. 

To promote the safe, reliable, and beneficial development of MLLMs, we propose the first hallucination attack utilizing only the attention sink phenomenon, to explore the current state of MLLM hallucination severity and existing mitigation strategies. By manipulating the attention mechanism and hidden states during generation, our method achieves dynamic, effective, and highly transferrable attacks, all without the need for additional human intervention. The attacking strategy requires no pre-defined target responses, and applies to general visual and textual inputs, overcoming a major limitation of current adversarial attacks against MLLMs.

\subsection{Security Settings}
\noindent \textbf{Attack Scenario.} 
We define the goal of our hallucination attack against MLLMs as increasing the amount of hallucinated content in MLLM responses, which includes inconsistencies in objects, attributes, and relationships according to the actual image content. The target models include open-sourced MLLMs and commercial MLLM APIs (such as GPT-4o provided by OpenAI). The target MLLMs may have built-in filtering mechanisms for multi-modal inputs or may be enhanced with additional mitigation strategies. The ultimate malicious goal can manifest in various forms, which include intensifying hallucination in subsequent conversations, delivering misleading information to users, and causing incorrect decisions in downstream applications, particularly in critical fields where even minor errors can have severe consequences. In real-world scenarios, many MLLM applications (e.g., AI assistants, customer service, physical therapy, and document analysis \cite{ye2023ureader}) allow user-defined uploads, which are vulnerable to adversarial inputs. Additionally, MLLMs in black-box systems (e.g., autonomous driving) are also susceptible to threats like physical adversarial patches \cite{hu2021naturalistic}.

\noindent \textbf{Threat Model.} 
We conceive an attacker who exploits adversarial visual inputs to perform hallucination attacks. With white-box access to a surrogate MLLM, the attacker can compute gradients and construct adversarial examples. The attacker's objective is to compromise the faithfulness of target MLLMs, thereby maliciously influencing their downstream applications. Driven by this goal, the attacker is highly motivated to deploy these adversarial inputs against black-box MLLMs and proprietary commercial MLLM APIs. To evade detection and filtering mechanisms employed by commercial platforms, the attacker must ensure that the adversarial inputs still produce high-quality and useful MLLM responses.

\subsection{Method}
\subsubsection{Formulation of MLLM Generation}
When processing multi-modal instructions during user interaction, the MLLMs take both visual and textual prompts as context. We denote the visual input tokens as $\textbf{x}^v = \{x_0, x_1, ..., x_{N-1}\}$, where $N$ represents the length of visual inputs, and is typically predefined during training. The textual input tokens are denoted as $\textbf{x}^t = \{x_N, x_{N+1}, ..., x_{M+N-1}\}$, with an input length of $M$. The multi-modal inputs are concatenated into a single sequence $\textbf{x}^{in} = \{x_i\}_{i=0}^{M+N-1}$, which is then fed into the MLLMs for further alignment and generation.

The response of MLLMs is sampled in an auto-regressive manner, with each token predicted based on previously generated ones. During the inference, the hidden states of each token are extracted as embeddings. We represent the hidden states of token $i$ in the $l$-th layer as $h_i^{(l)}$, and the self-attention scores of each token as $a_i^{(l)}$, which is formulated as:
\begin{equation}
    \textbf{h}^{(l)} = \{ h_0^{(l)}, h_1^{(l)},..., h_{T-1}^{(l)}\} \in \mathbb{R}^{T \times d}
\label{eq:hidden}
\vspace{-1em}
\end{equation}

\begin{equation}
    a_i^{(l)} = {\rm Softmax} \left( \frac{Q^{(l)} K^{{(l)}^\mathrm{T}}}{\sqrt{d_k}} \right)_i
\label{eq:attention}
\end{equation}

\begin{equation}
    \textbf{a}^{(l)} = \{ a_0^{(l)}, a_1^{(l)},..., a_{T-1}^{(l)}\} \in \mathbb{R}^{T \times T}
\end{equation}
where $T$ denotes the overall length of tokens including model responses, $d$ denotes the dimension of hidden states, $Q^{(l)}=\textbf{h}^{(l)} W_Q^{(l)} \in \mathbb{R}^{T \times d_k}$ and $K^{(l)}=\textbf{h}^{(l)} W_K^{(l)} \in \mathbb{R}^{T \times d_k}$ represents query and key vectors with a dimension of $d_k$, after the linear projection of hidden states. In the last layer of MLLMs, a vocabulary head $\mathcal{H}$ projects the final hidden states $\textbf{h}^{(L)}$ into probabilities, which supports the next-token prediction. 
\begin{equation}
    x_{<t} = \{x_0, x_1, ..., x_{t-1}\} 
\vspace{-0.8em}
\end{equation}

\begin{equation}
    p(x_t|x_{<t}) = {\rm Softmax} (\mathcal{H}(\textbf{h}^{(L)}))_{x_t}
\end{equation}
where $M+N \le t<T$, and $x_t \in \mathcal{X}$ in which $\mathcal{X}$ means the whole vocabulary set. With the probability of the next token $p(x_t|x_{<t})$, different decoding strategies such as beam search, greedy search, and nucleus sampling are adopted to output the model responses, which are further illustrated in Appendix \ref{sec:A1}. After predicting the $t$-th token, it will be appended at the end of the token list for the next-round generation. This process continues until the model predicts an end-of-sentence (EOS) token, signaling the end of responses. 

\subsubsection{Identifying Potential Sink Tokens}
According to our analysis in Section \ref{sec:3}, the attention sink appears at the turning point of image-text relevance, which aggregates some misleading global information from multi-modal inputs, and provokes hallucinated content.

When conducting the hallucination attack with adversarial visual inputs, we aim to coax the target MLLMs into generating as many attention sinks during their responses, which has been demonstrated to significantly increase the hallucinated content. Given that the maximum length of model responses and the decoding strategies cannot be fully covered in the attacking process, the adversarial perturbations are demanded to include the generalized patterns that trigger more sink tokens in MLLMs generation, rather than merely producing sink tokens at fixed token positions.

To identify the potential tokens that are most likely to become attention sinks, we perform a search in current model responses, based on the amount of global information contained in each token. We retrieve the hidden states of tokens $\textbf{h}^{(l)}$ in the intermediate layer of MLLMs, and compute the cosine similarity between the global input tokens $\textbf{x}^{in}$ and each generated token $x_i$ as follows:

\vspace{-0.5em}
\begin{equation}
    s_i^{(l)} = {\rm Sim}(h_i^{(l)}, \overline{h}_g^{(l)}), \ 1 \le l < L, \ i>M+N-1
\end{equation}
where ${\rm Sim}(\cdot, \cdot)$ represents the computation of cosine similarity, $\overline{h}_g^{(l)}$ is defined as the mean vector of the $l$-th layer hidden states of multi-modal inputs, i.e., $\overline{h}_g^{(l)} = {\rm Mean}(\{h_i^{(l)}\}_{i=0}^{M+N-1})$. 

Based on the global information each token absorbs within the middle-layer embeddings, we choose the token index with the highest $s_i^{(l)}$ as the potential sink position in the subsequent optimization of adversarial perturbation: 
\begin{equation}
    {\rm idx} = \mathcal{I} [\max \{ s_{M+N}^{(l)}, s_{M+N+1}^{(l)}, ..., s_{T-1}^{(l)}\}]
\end{equation}
where $\mathcal{I}[\cdot]$ indicates the index of the token with the maximum similarity score. 

In each round of adversarial optimization, we will identify the potential token according to current model responses. As the sentence structure remains stable in consecutive attack rounds, the selection of potential tokens tends to be focused over a few iterations, leading to concentrated and targeted optimization towards sink tokens. 

\subsubsection{Optimizing Adversarial Perturbations}
After searching for the potential token to induce attention sink, we design the adversarial target based on the ideal characteristics of sink tokens, which covers both attention behavior and knowledge aggregation. Introducing sink tokens with high attention scores, we demand that the subsequent generation includes declined image-text relevance, and produce hallucinated content from misleading information.

\noindent \textbf{Attention Loss.} 
When constructing the desired columnar attention behavior, we acquire the middle-layer attention scores within MLLM generation, which is $\textbf{a}^{(l)} = \{a_i^{(l)}\}_{i=0}^{T-1}$. The attention score of each token $a_i^{(l)} \in \mathbb{R}^i, 0 \le i < T$ contains its allocated attention on the previous sequence. We construct the attention map $A^{(l)}=[a_0^{(l)}, a_1^{(l)}, ..., a_{T-1}^{(l)}] \in \mathbb{R}^{T \times T}$ with obtained scores, where $a_{i,j}^{(l)}$ represents the attention scores of token $x_i$ allocated to $x_j$.

In the attention map $A^{(l)}$, a columnar attention pattern on certain tokens $x_i$ means that the subsequent tokens $x_j, j > i$ all allocate high attention to the sink token, and make predictions with its domination. To induce such a columnar attention pattern on the chosen potential token $x_{\rm idx}$, we focus on a localized attention window $A'^{(l)}=[a_{\rm idx}^{(l)}, a_{{\rm idx}+1}^{(l)}, ..., a_{T-1}^{(l)}]$, and the attention loss is defined as:
\begin{equation}
\label{eq:l_attn}
    \mathcal{L}_{\rm attn}(\textbf{x}^v, \textbf{x}^t) = CE(A'^{(l)}, {\rm idx})
\end{equation}
where $CE(\cdot)$ denotes the cross entropy loss function. The attention loss forces subsequent tokens to allocate high attention to the potential sink token, thereby forming the attention sink adversarially to introduce a decline of image-text relevance.

\noindent \textbf{Embedding Loss.} 
Besides the direct phenomenon of columnar attention behaviors, our observation also suggests that sink tokens bear a much higher resemblance to the multi-modal inputs, which aggregate misleading global information originating from the instruction tuning. In the hallucination attack, we aim to increase the cosine similarity of the embeddings between potential sink tokens and global input information. The hidden states of potential token in the $l$-th intermediate layer $h_{\rm idx}^{(l)}$ is obtained, and is used to compute its similarity of global information ${\rm Sim}(h_{\rm idx}^{(l)}, \overline{h}_g^{(l)})$. 

Since the aggregation of global information encourages the formation of an attention sink, we also introduce an embedding loss to partially raise the global information embedded in the potential tokens. We adopt the hinge loss to ensure the embedding only contains incomplete and misleading information:

\begin{equation}
\label{eq:l_emb}
    \mathcal{L}_{\rm emb}(\textbf{x}^v, \textbf{x}^t) = \max(0, \sigma - {\rm Sim}(h_{\rm idx}^{(l)}, \overline{h}_g^{(l)})) 
\end{equation}
where $\sigma$ is a hyper-parameter predefined with the observation of sink tokens in different MLLMs. The embedding loss is designed to promote the injection of more misleading global information into the potential token, which helps MLLMs spontaneously generate hallucinatory content in subsequent outputs. 

\noindent \textbf{Adversarial Objective.}
With the hallucination attack targeting both attention behavior and hidden states, the overall adversarial objective is defined as:

\begin{equation}
\begin{aligned}
    \min \ &\mathcal{L}_{\rm attn}(\tilde{\textbf{x}}^v, \textbf{x}^t) + \alpha \mathcal{L}_{\rm emb}(\tilde{\textbf{x}}^v, \textbf{x}^t) \\
    &{\rm s.t.}, \ \tilde{\textbf{x}}^v = \textbf{x}^v + \delta, \ ||\delta||_p < \epsilon
\end{aligned}
\end{equation}
where $\delta$ is the adversarial perturbation on the visual input $\textbf{x}^v$, $\alpha$ is the hyper-parameter to adjust the regularization, $||\cdot||_p$ is the computation of p-norm, and $\epsilon$ is the attack budget which controls the magnitude of the adversarial perturbation. 

The hallucination attack will be conducted in multiple rounds, where the adversarial objective will direct the optimization of adversarial perturbation in each round. The attacking algorithm of our method is illustrated in Alg. \ref{alg:atk}.

\begin{algorithm}
\setstretch{1.0}
\caption{Hallucination Attack}
\label{alg:atk}
\begin{algorithmic}[1]
\Statex \textbf{Input:} The visual input: $\mathbf{x}^v$, the textual input: $\mathbf{x}^t$, and the target model: \textit{mllm}

\Statex \textbf{Output:} The adversarial visual input: $\tilde{\mathbf{x}}^v$
\State $S \gets 0$.
\While{$S < S_{\rm max}$}
\State $response \gets \textit{mllm}(\tilde{\mathbf{x}}^v, \mathbf{x}^t)$.
\State Get hidden states $\textbf{h}^{(l)}$ from the $l$-th layer.
\State Get attention scores $\textbf{a}^{(l)}$ from the $l$-th layer.
\State $\overline{h}_g^{(l)} \gets {\rm Mean}(h^{(l)}_{0}, h^{(l)}_{1}, ..., h^{(l)}_{M+N-1})$.
\State Compute the cosine similarity between $\overline{h}_g^{(l)}$ and generated token embeddings as $s_i^{(l)} = {\rm Sim}(h_i^{(l)}, \overline{h}_g^{(l)})$
\State Get potential token ${\rm idx} = \mathcal{I} [\max \{ s_{M+N}^{(l)}, ..., s_{T-1}^{(l)}\}]$.
\State $A^{(l)} \gets$ construct attention map with $\textbf{a}^{(l)}$.
\State $\mathcal{L}_{\rm attn} \gets CE(A'^{(l)}, {\rm idx})$. \Comment{See Equation \ref{eq:l_attn}}
\State $\mathcal{L}_{\rm emb} \gets \max(0, \sigma - {\rm Sim}(h_{\rm idx}^{(l)}, \overline{h}_g^{(l)}))$.\Comment{See Equation \ref{eq:l_emb}}
\State Compute gradient $g \gets \nabla_{\tilde{\mathbf{x}}^v} \mathcal{L}_{\rm attn}+\alpha \mathcal{L}_{\rm emb}$
\State Updating $\tilde{\mathbf{x}}^v_{S} \gets \tilde{\mathbf{x}}^v_{S} - \gamma \cdot {\rm sign}(g)$.
\State Clipping $\tilde{\mathbf{x}}^v_{S} \gets {\rm Clip}(\tilde{\mathbf{x}}^v_{S}, -\epsilon, \epsilon)$.
\State $S \gets S + 1$
\EndWhile
\State \Return $\tilde{\mathbf{x}}^v$
\end{algorithmic}
\end{algorithm}
\vspace{-1.2em}
\section{Experiments}
\subsection{Experimental Settings}
\label{sec:5.1}

\begin{table*}[h]
\setlength{\belowcaptionskip}{0.15cm}
\caption{Basic information of open-source MLLMs in our experiments.}
\label{tab:model}
    \centering
    \begin{adjustbox}{width=0.8\textwidth}
    \begin{tabular}{ccccc}
    \toprule[1.5pt]
        \makebox[0.15\textwidth][c]{\textbf{MLLM}} & \makebox[0.15\textwidth][c]{\textbf{InstructBLIP}} & \makebox[0.15\textwidth][c]{\textbf{MiniGPT-4}} & \makebox[0.15\textwidth][c]{\textbf{LLaVA-1.5}} & \makebox[0.15\textwidth][c]{\textbf{Shikra}}  \\ \midrule
        \textbf{Visual Encoder} & Vicuna-7b-v1.1 & Vicuna-7b-v0 & Vicuna-7b & LLaMA-7b  \\
        \textbf{LLM Backbone} & EVA-ViT-g/14 & EVA-ViT-g/14 & CLIP-ViT-L/14 & CLIP-ViT-L/14  \\ 
        \bottomrule[1.5pt]
    \end{tabular}
\vspace{-0.8em}
\end{adjustbox}
\end{table*}

\noindent \textbf{Target Models. }
To conduct a comprehensive evaluation of mainstream MLLMs in the open-source community, we select four of the most representative MLLMs including InstructBLIP \cite{Dai2023InstructBLIPTG}, MiniGPT-4 \cite{zhu2023minigpt}, LLaVA-1.5 \cite{liu2024improved}, and Shikra \cite{Chen2023ShikraUM} as target models. The details about visual encoders and LLM backbones are available in Tab. \ref{tab:model}, which includes well-trained vision models like EVA \cite{fang2023eva} and CLIP \cite{radford2021learning}, and widely-used LLMs like Vicuna \cite{chiang2023vicuna} and LLaMA \cite{touvron2023llama}. To evaluate the adversarial effects on closed-source commercial APIs, we also take the recently released GPT-4o mini \cite{gpt4o} and Gemini 1.5 flash \cite{gemini} into consideration. More implementation details are available in Appendix \ref{sec:A2}.

\noindent \textbf{MLLM Tasks. } To comprehensively evaluate the adversarial impact of the hallucination attack, we focus on two types of downstream tasks for MLLMs: image captioning and question-answering (QA). These tasks represent different aspects of MLLMs' open-ended generation capabilities.  

\noindent \textbf{Metrics. }
We assume that the attacker's goal is to intensify hallucinations in MLLM responses without sacrificing their quality and helpfulness. For the image captioning task, we evaluate both the extent of hallucination and the quality of the generated responses. For the QA task, we evaluate the accuracy of model answers.

\noindent \textit{Evaluation of hallucination}. 
In assessing the severity of object hallucination, earlier research adopted the Caption Hallucination Assessment with Image Relevance (CHAIR) metric \cite{rohrbach2018object}. However, this metric fails to consider the hallucination of attributes and relationships, and only supports closed-ended evaluation (i.e., covering only 80 object classes in MS-COCO \cite{lin2014microsoft} datasets). To achieve more advanced evaluation for open-ended model responses, we follow the previous studies \cite{huang2024opera, liu2023mitigating} and conduct GPT-4 assisted evaluation based on HalluBench \cite{zhao2023beyond}. The benchmark is based on a subset of the Visual Genome (VG) dataset \cite{krishna2017visual}, featuring images annotated with bounding box coordinates and detailed descriptions of objects, attributes, and relationships. The HalluBench also provides human-annotated factual information to improve evaluation accuracy. During the evaluation, the GPT-4 model is prompted with all the references and judges the hallucination in MLLM responses sentence by sentence. The GPT-4 hallucination evaluation prompt is displayed in Fig. \ref{fig:hal_prompt}.

\noindent \textit{Evaluation of response quality}.
The overall quality of the MLLM's responses is critical in real-world attacking scenarios. If the adversarial visual inputs result in low-quality responses characterized by poor fluency, distorted sentence structure, or excessive grammatical errors, downstream users of MLLMs can easily detect these patterns and implement filtering mechanisms. In the evaluation, we also resort to the GPT-4 model to assess the grammar, fluency, and naturalness of generated responses. The model provides a score between 0-9 as the overall response quality. The GPT-4 assisted response quality evaluation prompt is displayed in Fig. \ref{fig:qua_prompt}.

\noindent \textit{Accuracy of QA}. To evaluate the attacking effects on the QA task, we use the OK-VQA \cite{marino2019ok} benchmark, which is based on the MS-COCO image dataset and widely recognized for assessing the general QA capabilities of MLLMs  \cite{Chen2023ShikraUM,bai2023qwen}. We select a subset of the OK-VQA benchmark, consisting of questions paired with 5 human-annotated answers each. For evaluation, we employ the standard VQA accuracy metric \cite{antol2015vqa}, which measures model responses by comparing them to ground truth answers. A response is considered 100\% accurate if at least three human annotators provide the exact same answer. The VQA accuracy metric is defined as:
\begin{equation}
    {\rm accuracy} = \min (\frac{{ \rm \#\ humans\ that\ provided\ that\ answer}}{3}, 1)
\label{eq:vqa_acc}
\end{equation}

\subsection{Adversarial Effects}
\label{sec:5.2}
\noindent \textbf{Image Captioning Task.}
To investigate the impact of adversarial visual inputs, we first concentrate on the level of hallucination in white-box models. During the hallucination attack, we construct adversarial visual inputs with the attack budget $\epsilon$ of $2/255$, $5/255$, and $8/255$, to observe the adversarial effects under different strengths of perturbations. During evaluation, we follow the setting of HalluBench, and query all MLLMs with a text prompt of \textit{Please describe this image in detail.}, together with the constructed adversarial visual inputs. 

The adversarial effects on white-box target MLLMs with beam search decoding are reported in Tab. \ref{tab:white_b}. Remarkably, our hallucination attack induces a substantial amount of hallucinated content in MLLM responses, achieving up to 75.74\% hallucinated words in single responses with only minor perturbations on visual inputs. With larger attack budgets, the number of affected sentences and words has noticeably increased. In the real-world applications of MLLMs, the severer hallucination in model responses may mislead downstream modules to focus on inaccurate descriptions of visual content and crash on making planning or decisions. It is also worth noting that, our hallucination attack does not work by increasing the output length (with similar SPI and WPI in results). This contrasts with the approach in \cite{gaoinducing} that delaying the occurrence of (EOS) token may lead to an increase in CHAIR metrics. However, their eight times longer model responses are easy to be detected from normal uses, and inevitably affect their fluency and helpfulness. More evaluation results of greedy search and nucleus sampling decoding are depicted in Tab. \ref{tab:white_g} and \ref{tab:white_n} in the appendix, also demonstrating remarkable adversarial effects.

\begin{table*}[]
\setlength{\belowcaptionskip}{0.15cm}
\caption{Results of GPT-4 assisted hallucination evaluation for the image captioning task on white-box models. All of the MLLM responses are generated with \textit{beam search} decoding. We report six aspects of evaluation, including the number of sentences per image (\textbf{SPI}), the number of words per image (\textbf{WPI}), the number of hallucinated sentences per image (\textbf{HSPI}), the number of hallucinated words per image (\textbf{HWPI}), the ratio of hallucinated sentences (\textbf{HSR}), and the ratio of hallucinated words (\textbf{HWR}). A larger HSPI, HWPI, HSR, and HWR indicate a higher level of hallucination in MLLM responses. The best results are marked in bold, and the number in brackets indicates the hallucination improvement compared to the clean image.}
\label{tab:white_b}
\newcommand{\improv}[1]{\footnotesize\textcolor[RGB]{17,74,133}{\textbf{(+#1)}}}
\newcommand{\reduce}[1]{\footnotesize\textcolor[RGB]{81,69,73}{\textbf{(-#1)}}}
\begin{adjustbox}{width=\textwidth}
\begin{tabular}{ccllllll}

\toprule[1.5pt]
\makebox[0.15\textwidth][c]{\textbf{Target Model}}   &
\makebox[0.11\textwidth][c]{\textbf{Input}} &
\makebox[0.08\textwidth][l]{\textbf{SPI}} & 
\makebox[0.08\textwidth][l]{\textbf{WPI}} & 
\makebox[0.1\textwidth][l]{\textbf{HSPI}} & 
\makebox[0.1\textwidth][l]{\textbf{HWPI}} & 
\makebox[0.1\textwidth][l]{\textbf{HSR(\%)}} & 
\makebox[0.1\textwidth][l]{\textbf{HWR(\%)}} \\
\midrule

\multirow{4}{*}{\textbf{InstructBLIP}} & clean image   & 4.54  & 75.64  & 2.83  & 48.05 & 62.91\% & 64.93\% \\
& $\epsilon$=2/255 & 4.60  & 80.19 & 2.97 \improv{0.14}& 55.14 \improv{7.09} & 64.92\% \improv{2.01\%} & 68.23\% \improv{3.30\%}\\
& $\epsilon$=5/255 & 4.47 & 80.48 & 3.04 \improv{0.21} & 54.90 \improv{6.85} & \textbf{68.41\%} \improv{5.50\%} & \textbf{70.84\%} \improv{5.91\%} \\
& $\epsilon$=8/255 & 4.41 & 79.71 & 2.89 \improv{0.06} & 52.91 \improv{4.86} & 66.79\% \improv{3.88\%} & 69.45\% \improv{4.52\%} \\

\midrule
\multirow{4}{*}{\textbf{LLaVA-1.5}}    & clean image   & 4.60 & 116.24 & 2.68 & 79.08  & 59.62\% & 71.68\%  \\
   & $\epsilon$=2/255 &4.64  &96.60  &2.76 \improv{0.08} &62.97 \reduce{16.11} &60.26\% \improv{0.64\%} &68.17\% \reduce{3.51\%} \\
   & $\epsilon$=5/255 &4.49  &108.03  &2.67 \reduce{0.01} &74.85 \reduce{4.23} &62.36\% \improv{2.74\%} &\textbf{75.74\%} \improv{4.06\%}  \\
   & $\epsilon$=8/255 &4.53  &103.58  &2.92 \improv{0.24} &75.45 \reduce{3.63} &\textbf{65.07\%} \improv{5.45\%} &75.08\% \improv{3.40\%} \\

\midrule
\multirow{4}{*}{\textbf{MiniGPT-4}}    & clean image   & 3.98 & 60.56 & 2.31 & 37.34 & 58.13\% &62.77\%  \\
   & $\epsilon$=2/255 & 4.10  & 59.20 & 2.49 \improv{0.18} & 37.97 \improv{0.63}& 61.42\% \improv{3.29\%}& 65.01\% \improv{2.24\%}\\
   & $\epsilon$=5/255 & 3.97 & 66.27 & 2.41 \improv{0.10}&  43.48 \improv{6.14} & 61.02\% \improv{2.89\%}& 67.09\% \improv{4.32\%}\\
   & $\epsilon$=8/255 & 4.00 & 64.51 & 2.55 \improv{0.24} & 40.83 \improv{3.49} & \textbf{64.59\%} \improv{6.46\%} & \textbf{67.97\%} \improv{5.20\%}\\

\midrule
\multirow{4}{*}{\textbf{Shikra}}   & clean image   & 3.11  & 46.13  &1.56  & 23.39 &52.95\%  &53.16\%  \\
   & $\epsilon$=2/255 & 3.13  &45.99  & 1.69 \improv{0.13} & 25.65 \improv{2.26}& 56.04\% \improv{3.09\%}& 57.93\% \improv{4.77\%}\\
   & $\epsilon$=5/255 &3.26 & 46.82 & 1.83 \improv{0.27} & 26.51 \improv{3.12}& \textbf{57.88\%} \improv{4.93\%} & 58.25\% \improv{5.09\%}\\
   & $\epsilon$=8/255 &3.12  &45.19  &1.69 \improv{0.13} &25.69 \improv{2.30} & 56.31\% \improv{3.36\%}& \textbf{59.11\%} \improv{5.95\%} \\
\bottomrule[1.5pt]  
\end{tabular}
\end{adjustbox}
\end{table*}

\begin{table*}[]
\setlength{\belowcaptionskip}{0.15cm}
\caption{Results of GPT-4 assisted hallucination evaluation for the image captioning task on black-box models. All of the MLLM responses are generated with \textit{beam search} decoding. The six aspects of evaluation are the same as in Tab. \ref{tab:white_b}. A larger HSPI, HWPI, HSR, and HWR indicate a higher level of hallucination in MLLM responses. The best results are marked in bold, and the number in brackets indicates the hallucination improvement compared to the clean image for each target model.}
\label{tab:black_b}
\newcommand{\improv}[1]{\footnotesize\textcolor[RGB]{17,74,133}{\textbf{(+#1)}}}
\newcommand{\reduce}[1]{\footnotesize\textcolor[RGB]{81,69,73}{\textbf{(-#1)}}}
\begin{adjustbox}{width=\textwidth}
\begin{tabular}{ccllllll}

\toprule[1.5pt]
\makebox[0.15\textwidth][c]{\textbf{Surrogate Model}}   &
\makebox[0.11\textwidth][c]{\textbf{Target Model}} &
\makebox[0.08\textwidth][l]{\textbf{SPI}} & 
\makebox[0.08\textwidth][l]{\textbf{WPI}} & 
\makebox[0.1\textwidth][l]{\textbf{HSPI}} & 
\makebox[0.1\textwidth][l]{\textbf{HWPI}} & 
\makebox[0.1\textwidth][l]{\textbf{HSR(\%)}} & 
\makebox[0.1\textwidth][l]{\textbf{HWR(\%)}} \\
\midrule

\multirow{4}{*}{\textbf{InstructBLIP}} & InstructBLIP   &4.47  &80.48  &3.04 \improv{0.21} &54.90 \improv{6.85} &68.41\% \improv{5.50\%} &70.84\% \improv{5.91\%} \\
& LLaVA-1.5 &4.46  &99.77  &2.64 \reduce{0.04} &70.51 \reduce{8.57} &59.42\% \reduce{0.20\%} &71.48\% \reduce{0.20\%} \\
& MiniGPT-4 &3.84  &63.00  &2.31  &40.54 \improv{3.20} &\textbf{61.81\%} \improv{3.68\%} &\textbf{68.21\%} \improv{5.44\%} \\
& Shikra &3.20  &48.95  &1.79 \improv{0.23} &27.77 \improv{4.38} &56.14\% \improv{3.19\%} &57.09\% \improv{3.93\%} \\

\midrule
\multirow{4}{*}{\textbf{LLaVA-1.5}}    & LLaVA-1.5   &4.49  &108.03  &2.67 \reduce{0.01} &74.85 \reduce{4.23} &62.36\% \improv{2.74\%} &75.74\% \improv{4.06\%} \\
   & InstructBLIP &4.47  &78.31  &2.81 \reduce{0.02} &51.85 \improv{3.80} &65.37\% \improv{2.46\%} &68.75\% \improv{3.82\%} \\
   & MiniGPT-4 &3.95  &63.60  &2.32 \improv{0.01} &42.25 \improv{4.91} &60.79\% \improv{2.66\%} &68.14\% \improv{5.37\%} \\
   & Shikra &3.08  &45.94  &1.94 \improv{0.38} &29.96 \improv{6.57} &\textbf{63.85\%} \improv{10.90\%} &\textbf{65.90\%} \improv{12.74\%} \\

\midrule
\multirow{4}{*}{\textbf{MiniGPT-4}}    & MiniGPT-4   & 4.00 & 64.51 & 2.55 \improv{0.24} & 40.83 \improv{3.49} & 64.59\% \improv{6.46\%} & 67.97\% \improv{5.20\%}\\
   & InstructBLIP &4.36  &79.62  &2.96 \improv{0.13} &54.82 \improv{6.77} &\textbf{68.96\%} \improv{6.05\%} &\textbf{71.94\%} \improv{7.01\%} \\
   & LLaVA-1.5 &4.27  &116.50  &2.51 \reduce{0.17} &75.86 \reduce{3.22} &60.84\% \improv{1.22\%} &73.67\% \improv{1.99\%} \\
   & Shikra &3.33  &49.67  &1.92 \improv{0.36} &28.99 \improv{5.60} &58.86\% \improv{5.91\%} &59.61\% \improv{6.45\%} \\

\midrule
\multirow{4}{*}{\textbf{Shikra}}   
& Shikra   &3.12  &45.19  &1.69 \improv{0.13} &25.69 \improv{2.30} & 56.31\% \improv{3.36\%}& 59.11\% \improv{5.95\%} \\
   & InstructBLIP &4.48  &80.36  &2.90 \improv{0.07} &54.39 \improv{6.34} &67.48\% \improv{4.57\%} &70.28\% \improv{5.35\%} \\
   & LLaVA-1.5 &4.43  &110.61  &2.71 \improv{0.03} &75.34 \reduce{3.74} &\textbf{64.77\%} \improv{5.15\%} &77.35\% \improv{5.67\%} \\
   & MiniGPT-4 &3.97  &72.35  &2.35 \improv{0.04} &46.90 \improv{9.56} &60.86\% \improv{2.73\%} &\textbf{70.07\%} \improv{7.30\%} \\
\bottomrule[1.5pt]  
\end{tabular}
\end{adjustbox}
\end{table*}

\begin{table*}[h]
\setlength{\belowcaptionskip}{0.18cm}
\caption{Results of QA accuracy on the OK-VQA benchmark. All of the MLLM answers are generated with \textit{greedy} decoding. A lower accuracy indicates a higher level of hallucination in MLLM answers. The best results are marked in bold, and the number in brackets indicates the hallucination improvement compared to the clean image.}
\label{tab:qa}
\newcommand{\improv}[1]{\footnotesize\textcolor[RGB]{17,74,133}{\textbf{(-#1)}}}
\newcommand{\reduce}[1]{\footnotesize\textcolor[RGB]{81,69,73}{\textbf{(+#1)}}}

\centering
\begin{adjustbox}{width=\textwidth}
\begin{tabular}{ccllll}

\toprule[1.5pt]
\makebox[0.14\textwidth][c]{\textbf{Target Model}}   & \makebox[0.12\textwidth][c]{\textbf{Input}} & 
\multicolumn{4}{c}{\textbf{Surrogate Model}}  \\
\midrule
&    & \makebox[0.17\textwidth][l]{\textbf{InstructBLIP}} & \makebox[0.17\textwidth][l]{\textbf{LLaVA-1.5}} & \makebox[0.17\textwidth][l]{\textbf{MiniGPT-4}} & 
\makebox[0.17\textwidth][l]{\textbf{Shikra}} \\ 
\midrule
\multirow{4}{*}{\textbf{InstructBLIP}} & clean image & \multicolumn{4}{c}{56.33}\\
& $\epsilon$=2/255 & 51.33 \improv{5.00} & 55.99 \improv{0.34} & 52.33 \improv{4.00} & 56.66 \reduce{0.33} \\
& $\epsilon$=5/255 & 48.99 \improv{7.34} & 53.99 \improv{2.34} & 50.33 \improv{6.00} & 50.33 \improv{6.00} \\
& $\epsilon$=8/255 & 49.66 \improv{6.67} & 49.99 \improv{6.34} & \textbf{47.33} \improv{9.00} & 52.33 \improv{4.00} \\
\midrule
\multirow{4}{*}{\textbf{LLaVA-1.5}} & clean image  & \multicolumn{4}{c}{60.00}  \\
& $\epsilon$=2/255 & 56.66 \improv{3.34} & 57.33 \improv{2.67} & 57.66 \improv{2.34} & 56.99 \improv{3.01} \\
& $\epsilon$=5/255 & 55.99 \improv{4.01} & 56.66 \improv{3.34} & 56.33 \improv{3.67} & \textbf{52.33} \improv{7.67} \\
& $\epsilon$=8/255 & 54.66 \improv{5.34} & \textbf{52.33} \improv{7.67} & 54.99 \improv{5.01} & 58.66 \improv{1.34} \\
\midrule
\multirow{4}{*}{\textbf{MiniGPT-4}} & clean image  & \multicolumn{4}{c}{40.66}  \\
& $\epsilon$=2/255 & 42.33 \reduce{1.67} & 38.33 \improv{2.33} & 39.33 \improv{1.33} & 42.33 \reduce{1.67} \\
& $\epsilon$=5/255 & 39.33 \improv{1.33} & 43.66 \reduce{3.00} & 36.66 \improv{4.00} & 40.33 \improv{0.33} \\
& $\epsilon$=8/255 & 39.33 \improv{1.33} & 43.33 \reduce{2.67} & \textbf{34.00} \improv{6.66} & 39.99 \improv{0.67} \\
\midrule
\multirow{4}{*}{\textbf{Shikra}} & clean image  & \multicolumn{4}{c}{55.33}  \\
& $\epsilon$=2/255 & 55.00 \improv{0.33} & 54.66 \improv{0.67} & 56.99 \reduce{1.66} & 56.66 \reduce{1.33} \\
& $\epsilon$=5/255 & 56.66 \reduce{1.33} & 54.00 \improv{1.33} & 53.33 \improv{2.00} & \textbf{51.33} \improv{4.00} \\
& $\epsilon$=8/255 & 51.66 \improv{3.67} & 52.99 \improv{2.34} & 52.66 \improv{2.67} & 53.33 \improv{2.00} \\

\bottomrule[1.5pt]
\end{tabular}
\end{adjustbox}
\vspace{-0.3em}
\end{table*}

\begin{figure*}[h]
    \centering
\includegraphics[width=\textwidth]{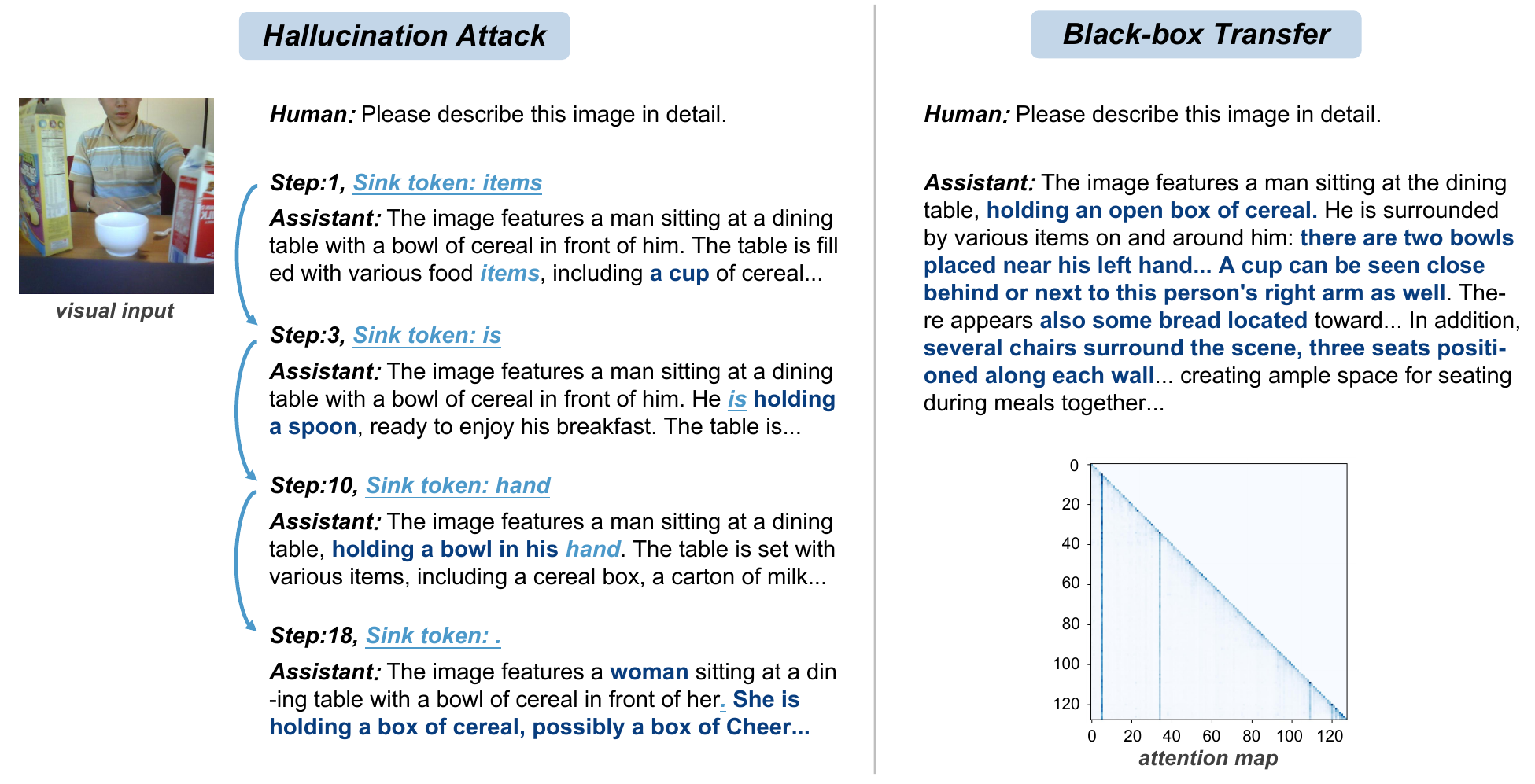}
\vspace{-2em}
\caption{\textbf{Left}: The visualization of sink tokens and model responses during the hallucination attack. We display an optimization process of LLaVA-1.5 on the HalluBench dataset. \textbf{Right}: The black-box transfer effect of the constructed adversarial visual input on InstructBLIP. A pronounced attention sink phenomenon is observed in the attention map. The attention sink in model responses is denoted with \textcolor[RGB]{76,153,202}{\textbf{blue tokens}}, and the hallucinated content is denoted with \textcolor[RGB]{8,61,126}{\textbf{indigo ones}}.}
\label{fig:loss}
\end{figure*}

During the attack process, we observe that as the adversarial noise was continuously optimized, changes in the sink token and MLLM responses align with our analysis in Section \ref{sec:3}. Fig. \ref{fig:loss} visualizes the attack process and the affected model response. During the optimization, the sink token changes from tokens with concrete meaning (e.g., \textit{item} in step 1) to non-content ones (e.g., the comma \textit{,} in step 18). The manipulation of attention behaviors also leads to more hallucinated objects (e.g., \textit{cup} and \textit{woman}) and wrong relationships (e.g., \textit{holding a bowl in his hand}) during the attack process. With adversarial visual inputs, the target model tends to make up contents not aligned with images, with an obvious attention sink phenomenon observed in the attention map. 

\noindent \textbf{QA Task.}
The white-box adversarial effects on the OK-VQA benchmark is presented in Tab. \ref{tab:qa}. The results show a reduction in accuracy of up to 7.67\%, emphasizing that adversarial visual inputs significantly impair the general visual understanding capabilities of MLLMs. As a result, hallucinated answers may be generated and presented to downstream users, potentially leading to the propagation of misconceptions in real-world applications.

To gain a deeper understanding of the adversarial effect on model responses, we engaged human experts to classify the hallucinations induced by our attack. The hallucinations are categorized into 5 different types with varying severity, as presented in Fig. \ref{fig:halu_human}. For further illustration, we provide qualitative analyses with several cases shown in Fig. \ref{fig:case1} to Fig. \ref{fig:case4} in the appendix.

\subsection{Black-box Transferability}
\label{sec:5.3}
In real-world attack scenarios, the target MLLMs usually remain inaccessible to the attackers. To demonstrate the effectiveness of hallucination attack under such settings, we construct adversarial visual inputs on one surrogate MLLM, and evaluate the transferability to both black-box MLLMs and closed-source commercial APIs. The attack is more challenging since the model structure, parameters and training paradigms of target MLLMs are unknown to the attackers. Commercial API providers such as OpenAI may also implement advanced defenses against multi-modal inputs \cite{ying2024unveiling}, which have successfully blocked semantics-based visual attacks \cite{liu2023query} and typographic attacks \cite{gong2023figstep}.

\subsubsection{Attacking Black-box MLLMs.}
\label{sec:5.3.1}
\noindent \textbf{Image Captioning Task.}
In our experimental setup, we select one of the four target MLLMs in Section \ref{sec:5.1} as the surrogate model, and evaluate the black-box transferability on the remaining ones. The results of hallucination evaluation for the image captioning task are presented in Tab. \ref{tab:black_b}. Our proposed attack demonstrates high transferability across different structures of MLLMs, with the perturbed visual inputs achieving a 10.90\% HSR and 12.74\% HWR increase on black-box target models, even surpassing the increase in white-box attacks. We also find that the attack achieves better transferability on MLLMs with the same visual encoder architecture, due to the similar distribution of hidden states and the effects of disturbing the attention behaviors. More evaluation results of greedy search and nucleus sampling decoding are depicted in Tab. \ref{tab:black_g} and \ref{tab:black_n} in the appendix. 

\noindent \textbf{QA Task.}
The black-box evaluation results for the QA task are reported in Tab. \ref{tab:qa}, demonstrating strong transferability across different architectures. The occasional increases in the accuracy of adversarial visual inputs may be attributed to the limitations of the manually labeled answers. 

\subsubsection{Attacking Closed-source Commercial APIs.}
\label{sec:5.3.2}
\begin{figure}[h]
    \centering
\includegraphics[width=\columnwidth]{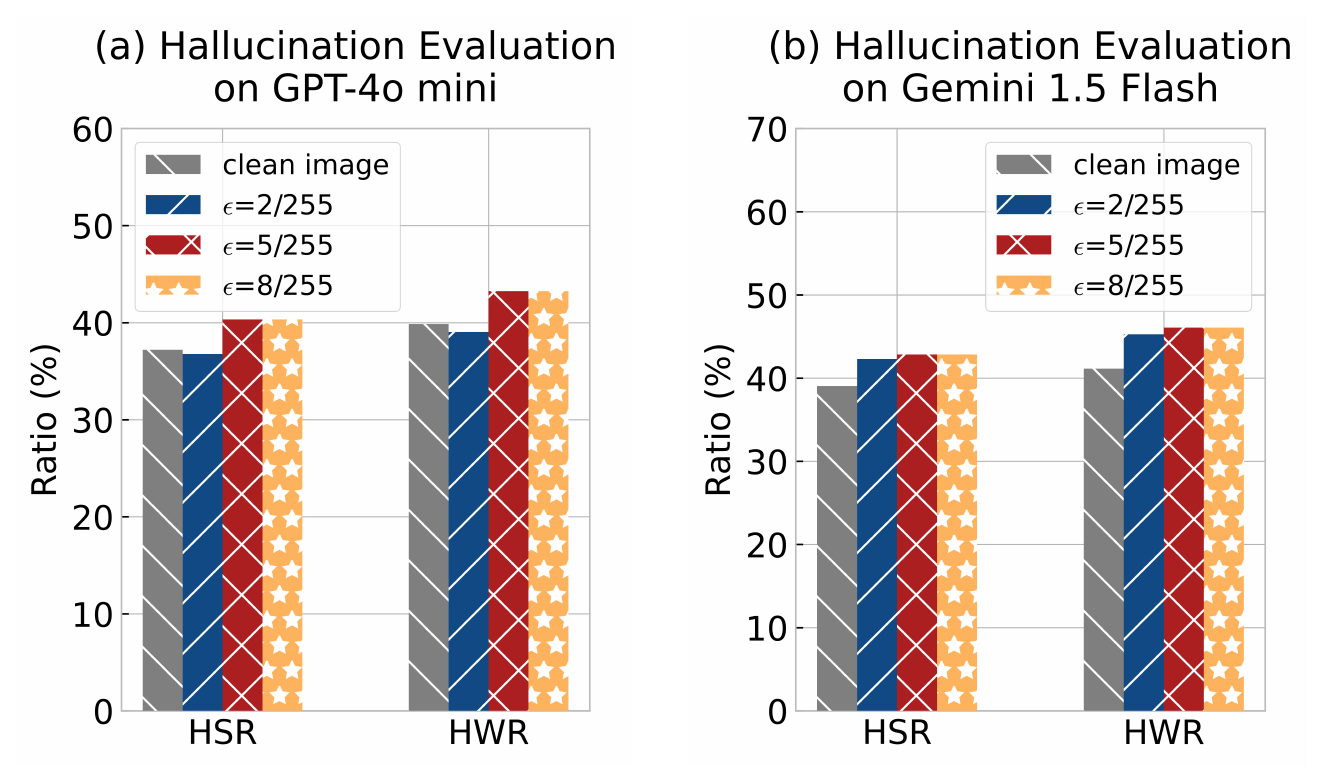}
\vspace{-1.8em}
\caption{Results of GPT-4 assisted hallucination evaluation for image captioning task on commercial APIs: \textbf{(a)} GPT-4o mini released by OpenAI and \textbf{(b)} Gemini 1.5 Flash launched by Google. A larger HSR and HWR indicate more hallucinations in MLLM responses.}
\label{fig:api}
\vspace{-1em}
\end{figure}

\noindent \textbf{Image Captioning Task.}
As for attacking commercial APIs, our goal is to examine whether the proposed attack could evade potential defense and detection measures implemented by IT giants. We choose the latest commercial MLLM APIs, i.e., the GPT-4o mini \cite{gpt4o} and Gemini 1.5 flash \cite{gemini}, as the target APIs. The adversarial transferability result for image captioning is displayed in Fig. \ref{fig:api}. Though potential defenses are applied, the proposed attack still achieves a 3.40\% and 5.32\% increase in hallucination words. As the captions of the commercial MLLM APIs have been utilized in applications like medical diagnosis \cite{wu2023can}, science education \cite{lee2023gemini}, and financial decisions \cite{lin2024harnessing}, the inaccurate interpretation induced by adversarial visual inputs deserves attention.

\noindent \textbf{QA Task.}
In the QA task, the adversarial effects in Fig. \ref{fig:api_qa} also reveal degraded visual comprehension and question-answering capabilities. Since commercial MLLM APIs are often regarded as powerful domain experts to guide the fine-tuning process (e.g., in medical \cite{sun2024self,liu2024gemex} and financial \cite{gan2024mme,xie2024open} domains), our attack may significantly undermine the reliability of domain-specific MLLMs. 

\subsection{The Quality of Model Responses}
\label{sec:5.4}
\begin{figure}[h]
    \centering
\includegraphics[width=0.98\columnwidth]{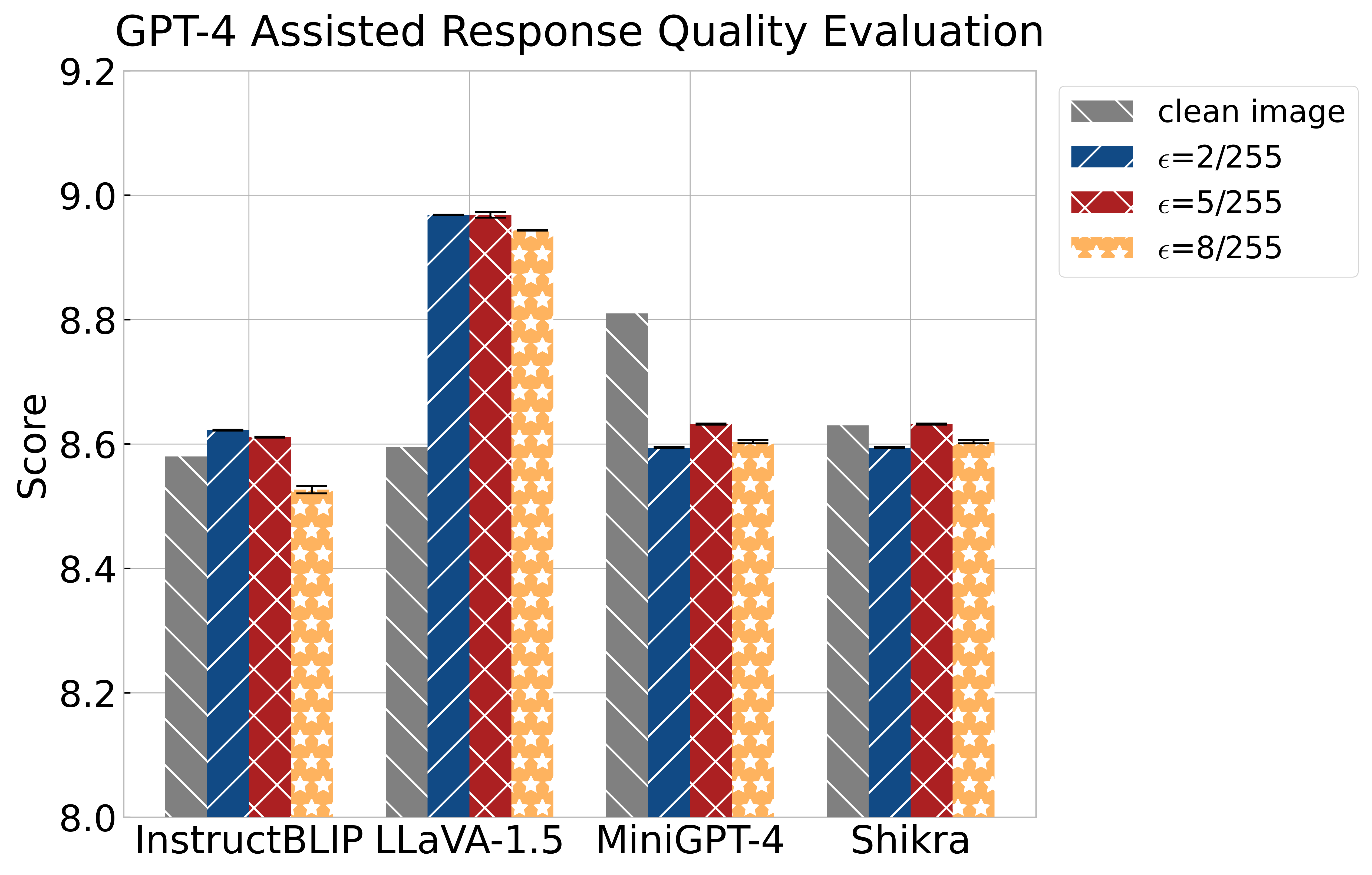}
\vspace{-0.5em}
\caption{Results of GPT-4 assisted response quality evaluation, covering both white-box and black-box attack scenarios. A higher score reflects better MLLM response quality.}
\label{fig:quality}
\end{figure}

When the attacker injects the adversarial visual input into target MLLMs, the model responses should resemble those in normal conversations, with no noticeable decline in quality. Otherwise, the stealth of the hallucination attack cannot be ensured. We report the results of response quality for the image captioning task based on GPT-4 assisted evaluation in Fig. \ref{fig:quality}, and the results based on the Perplexity metric (PPL) in Fig. \ref{fig:ppl}. Even if the model responses include more hallucinated content, they still maintain a high level of semantic accuracy, usefulness, and fluency, indicating the feasibility and stealthiness of our attack in practical settings.

\subsection{Attacking Mitigation Mechanisms}
As discussed in Section \ref{sec:2.2}, recent mitigation mechanisms of MLLM hallucination fall into three categories: mitigation with \textit{decoding}, \textit{model retraining}, and \textit{post-processing}. To verify the effectiveness of our attack on existing defenses, we select the representative methods as follows. All experiments are performed with the official implementation and released checkpoints in default parameters. 

\noindent \textbf{OPERA} \cite{huang2024opera}.
Targeting the attention sink phenomenon, this method detects the columnar behavior in MLLM's self-attention matrices, and applies penalty terms to the model logits during the beam search inference. As a decoding-based mitigation approach, it directly blocks the attention behaviors associated with hallucination attacks, establishing a strong baseline for mitigation.

\noindent \textbf{VCD} \cite{leng2024mitigating}.
Aiming to mitigate the inherent language prior, this method manipulates the decoding strategies of MLLMs, and adjusts the output logits with distorted visual inputs, which reflects the model's reliance on statistical bias. 

\noindent \textbf{Less is more} \cite{yue2024less}.
Noticing the overly detailed descriptions in MLLM instruction-tuning datasets, this method fine-tunes released MLLMs with additional supervision on EOS decisions, encouraging the models to stop generation timely before hallucinated content. 

\noindent \textbf{LRV-Instruction} \cite{liu2023mitigating}.
Addressing the limitation of existing instruction-tuning datasets that only contain samples of positive answers, this method constructs a comprehensive dataset with negative samples to guide models in identifying incorrect instructions. The dataset is used to fine-tune released MLLMs to mitigate hallucination.

\noindent \textbf{LURE} \cite{zhouanalyzing}.
With a post-hoc rectification strategy, this method collects a hallucinatory dataset and fine-tunes another MLLM as the hallucination revisor. During inference, the trained revisor detects underlying object hallucinations with output logits and rectifies them with new responses. 

\begin{table}[h]
\newcommand{\improv}[1]{\footnotesize\textcolor[RGB]{17,74,133}{\textbf{(+#1)}}}
\newcommand{\reduce}[1]{\footnotesize\textcolor[RGB]{81,69,73}{\textbf{(-#1)}}}

\setlength{\belowcaptionskip}{0.15cm}
\caption{Results of GPT-4 assisted hallucination evaluation against mitigation mechanisms on LLaVA-1.5. $(^{\ast})$, $(^{\circ})$, and $(^{\diamond})$ denote methods through \textit{decoding}, \textit{model retraining}, and \textit{post-processing} respectively. Best results are marked in bold.}
\label{tab:llava}
\begin{adjustbox}{width=\columnwidth}
\begin{tabular}{ccll}
\toprule[1.5pt]
\makebox[0.1\columnwidth][c]{\textbf{Mitigation}}  & 
\makebox[0.02\columnwidth][c]{\textbf{Input}} &  \makebox[0.2\columnwidth][l]{\textbf{HSR(\%)}} & 
\makebox[0.2\columnwidth][l]{\textbf{HWR(\%)}} \\
\midrule
\multirow{4}{*}{\parbox{2cm}{\centering \textbf{OPERA$^{\ast}$\\}\cite{huang2024opera}}}  & clean   &50.27\%    &51.93\%  \\
    & $\epsilon$=2/255 &53.50\% \improv{3.23\%}      &56.13\% \improv{4.20\%}   \\
    & $\epsilon$=5/255 &52.33\% \improv{2.06\%}     &54.37\% \improv{2.44\%}   \\
    & $\epsilon$=8/255 &\textbf{55.86\%} \improv{5.59\%}    &\textbf{58.18\%} \improv{6.25\%}   \\
\midrule
\multirow{4}{*}{\parbox{2cm}{\centering \textbf{VCD$^{\ast}$\\}\cite{leng2024mitigating}}}    & clean   &51.38\%      &53.58\%    \\
    & $\epsilon$=2/255 &54.46\% \improv{3.08\%}    &57.02\% \improv{3.44\%}   \\
    & $\epsilon$=5/255 &57.69\% \improv{6.31\%}    &60.12\% \improv{6.54\%}   \\
    & $\epsilon$=8/255 &\textbf{62.42\%} \improv{11.04\%}    &\textbf{64.95}\% \improv{11.37\%}   \\
\midrule
\multirow{4}{*}{\parbox{2cm}{\centering \textbf{Less is\\More$^{\circ}$} \cite{yue2024less}}} & clean   &43.74\%     &45.78\%    \\
    & $\epsilon$=2/255 &46.22\% \improv{2.48\%}    &48.23\% \improv{2.45\%}   \\
    & $\epsilon$=5/255 &47.68\% \improv{3.94\%}    &49.91\% \improv{4.13\%}  \\
    & $\epsilon$=8/255 &\textbf{52.77\%} \improv{9.03\%}    &\textbf{54.07\%} \improv{8.29\%}   \\

\bottomrule[1.5pt]  
\end{tabular}
\end{adjustbox}
\vspace{-1em}
\end{table}
\begin{table}[h]
\newcommand{\improv}[1]{\footnotesize\textcolor[RGB]{17,74,133}{\textbf{(+#1)}}}
\newcommand{\reduce}[1]{\footnotesize\textcolor[RGB]{81,69,73}{\textbf{(-#1)}}}

\setlength{\belowcaptionskip}{0.15cm}
\caption{Results of GPT-4 assisted hallucination evaluation against mitigation mechanisms on MiniGPT-4. $(^{\ast})$, $(^{\circ})$, and $(^{\diamond})$ denote methods through \textit{decoding}, \textit{model retraining}, and \textit{post-processing} respectively. Best results are marked in bold.}
\label{tab:minigpt4}
\begin{adjustbox}{width=\columnwidth}
\begin{tabular}{ccll}
\toprule[1.5pt]
\makebox[0.1\columnwidth][c]{\textbf{Mitigation}}  & 
\makebox[0.02\columnwidth][c]{\textbf{Input}} &  \makebox[0.2\columnwidth][l]{\textbf{HSR(\%)}} & 
\makebox[0.2\columnwidth][l]{\textbf{HWR(\%)}} \\
\midrule
\multirow{4}{*}{\parbox{2cm}{\centering \textbf{OPERA$^{\ast}$\\}\cite{huang2024opera}}}  & clean   &43.71\%     &45.79\%  \\
    & $\epsilon$=2/255 &57.09\% \improv{13.38\%}       &59.34\% \improv{13.55\%}   \\
    & $\epsilon$=5/255 &\textbf{60.78\%} \improv{17.07\%}     &\textbf{63.75\%} \improv{17.96\%}   \\
    & $\epsilon$=8/255 &59.03\% \improv{15.32\%}    &61.82\% \improv{16.03\%}   \\
\midrule
\multirow{4}{*}{\parbox{2cm}{\centering \textbf{LRV-\\Instruction$^{\circ}$} \cite{liu2023mitigating}}}    & clean   &67.19\%     &70.82\%    \\
    & $\epsilon$=2/255 &69.73\% \improv{2.54\%}    &73.81\% \improv{2.99\%}   \\
    & $\epsilon$=5/255 &70.75\% \improv{3.56\%}    &75.03\% \improv{4.21\%}   \\
    & $\epsilon$=8/255 &\textbf{71.43\%} \improv{4.24\%}    &\textbf{75.54\%} \improv{4.72\%}   \\
\midrule
\multirow{4}{*}{\parbox{2cm}{\centering \textbf{LURE$^{\diamond}$\\}\cite{zhouanalyzing}}} & clean   &48.57\%     &53.54\%    \\
    & $\epsilon$=2/255 &58.21\% \improv{9.64\%}    &64.44\% \improv{10.90\%}   \\
    & $\epsilon$=5/255 &59.42\% \improv{10.85\%}    &67.41\% \improv{13.87\%}   \\
    & $\epsilon$=8/255 &\textbf{59.97\%} \improv{11.40\%}    &\textbf{67.85\%} \improv{14.31\%}   \\

\bottomrule[1.5pt]  
\end{tabular}
\end{adjustbox}
\end{table}

\noindent \textbf{Attacking Mitigation Mechanisms.} 
\label{sec:5.5}
We assess the adversarial effects on LLaVA-1.5 and MiniGPT-4 with the representative mitigation methods, and the results are detailed in Tab. \ref{tab:llava} and Tab. \ref{tab:minigpt4} respectively. Experimental results demonstrate that the proposed attack succeeded in breaking all mitigation mechanisms in our evaluation, bringing the hallucination rate back to the level when no defenses are in place (e.g., the HWR achieves 75.54\% in attacking LRV-Instruction defense, outperforming the one on vanilla model). This indicates the efficacy of our attack in bypassing existing mitigation methods, including adaptive strategies like OPERA. It also reveals the shortcomings of current mitigation strategies in defending deliberately crafted perturbations.

\begin{table*}[!h]
\setlength{\belowcaptionskip}{0.15cm}
\caption{Results of GPT-4 assisted hallucination evaluation of the baseline method with \textit{beam search} decoding. $\delta$ denotes of budget of random noises injected into visual inputs. The line of \textit{attack} denotes the best results in the white-box attack scenario.}
\label{tab:baseline_b}
\newcommand{\improv}[1]{\scriptsize\textcolor[RGB]{17,74,133}{\textbf{(+#1)}}}
\newcommand{\reduce}[1]{\scriptsize\textcolor[RGB]{81,69,73}{\textbf{(-#1)}}}

\begin{adjustbox}{width=\textwidth}
\begin{tabular}{cllllllll}
\toprule[1.5pt]
\makebox[0.02\textwidth][c]{} & \multicolumn{2}{l}{\ \ \ \ \ \ \ \ \ \ \textbf{InstructBLIP}} & 
\multicolumn{2}{l}{\ \ \ \ \ \ \ \ \ \ \ \ \ \textbf{LLaVA-1.5}} & \multicolumn{2}{l}{\ \ \ \ \ \ \ \ \ \ \ \ \ \textbf{MiniGPT-4}}        & \multicolumn{2}{l}{\ \ \ \ \ \ \ \ \ \ \ \ \ \ \ \  \textbf{Shikra}}   \\
\midrule
& \makebox[0.02\textwidth][l]{\textbf{HSR(\%)}}    & \makebox[0.02\textwidth][l]{\textbf{HWR(\%)}}    & \makebox[0.02\textwidth][l]{\textbf{HSR(\%)}}    & \makebox[0.02\textwidth][l]{\textbf{HWR(\%)}}    & \makebox[0.02\textwidth][l]{\textbf{HSR(\%)}}    & \makebox[0.02\textwidth][l]{\textbf{HWR(\%)}}    & \makebox[0.02\textwidth][l]{\textbf{HSR(\%)}}    & \makebox[0.02\textwidth][l]{\textbf{HWR(\%)}}  \\
\midrule
clean & 62.91\% & 64.93\% & 59.62\% & 71.68\% & 58.13\% & 62.77\% & 52.95\% & 53.16\%  \\
$\delta$=2/255 & 60.34\% \reduce{2.57\%} & 63.14\% \reduce{1.79\%} & 57.37\% \reduce{2.25\%} & 73.14\% \improv{1.46\%} & 63.35\% \improv{5.22\%}  &     65.87\% \improv{3.10\%} & 56.07\% \improv{3.12\%} & 54.76\% \improv{1.60\%}  \\
$\delta$=5/255 & 63.27\% \improv{0.36\%} & 70.22\% \improv{5.29\%} & 59.12\% \reduce{0.50\%}       & 71.18\% \reduce{0.50\%} &  61.11\% \improv{2.98\%} & 63.24\% \improv{0.47\%} & 51.88\% \reduce{1.07\%}  &  52.95\% \reduce{0.21\%} \\
$\delta$=8/255 & 60.91\% \reduce{2.00\%} & 66.79\% \improv{1.86\%} & 57.39\% \reduce{2.23\%}     & 69.03\% \reduce{2.65\%} & 58.70\% \improv{0.57\%}  & 62.78\% \improv{0.01\%}  & 51.75\% \reduce{1.20\%}  & 52.81\% \reduce{0.35\%}  \\
\midrule
attack  & \textbf{68.41\%} \improv{5.50\%} & \textbf{70.84\%} \improv{5.91\%} & \textbf{65.07\%} \improv{5.45\%}  & \textbf{75.74\%} \improv{4.06\%}  & \textbf{64.59\%} \improv{6.46\%}   & \textbf{67.97\%} \improv{5.20\%}   & \textbf{57.88\%} \improv{4.93\%}   & \textbf{59.11\%} \improv{5.95\%}  \\
\bottomrule[1.5pt]
\end{tabular}
\end{adjustbox}
\vspace{-0.2em}
\end{table*}
\noindent \textbf{Attacking Adaptive Mitigation.} 
Since the emergence of sink tokens triggers hallucinated content, we consider an early-stopping adaptive mitigation. With white-box access to the target MLLMs, this mitigation detects the attention sink phenomenon during generation and terminates output before sink tokens appear. Tab. \ref{tab:adaptive} reports the length, quality, and hallucination metrics of MLLM responses under the mitigation strategy. Despite the implementation of adaptive mitigation, the adversarial visual inputs continue to provoke more severe hallucinations in model outputs. While adaptive mitigation reduces hallucinated content, it significantly decreases the mean length and quality of responses by 45.36\% and 63.67\% respectively, resulting in incomplete and less detailed descriptions of image content. In real-world applications, this strategy may also incur substantial computational overhead and severely degrade the user experience for commercial APIs.

\subsection{Baseline Comparison}
\label{sec:5.6}

To demonstrate the superiority of our attack in constructing adversarial perturbations that result in a severe level of hallucination, we consider visual inputs with random Gaussian noises as a baseline. The magnitude of random perturbation is set as the same of hallucination attacks. The attack effects of baseline methods with beam search decoding are displayed in Tab. \ref{tab:baseline_b}. It is obvious that trivial random perturbation, though disrupting the feature extraction process of the visual encoder and enhancing the model's reliance on language priors, shows no significant attack effect in hallucinated content. The comparison highlights the importance of manipulating the inherent behaviors of MLLMs to influence their generation process. More comparison results of greedy search and nucleus sampling decoding are available in Tab. \ref{tab:baseline_g} and \ref{tab:baseline_n} in the appendix.

\section{Discussions}
\noindent \textbf{Alignment of GPT-4 Assisted Evaluation.} To assess the alignment and stability of GPT-4-based metrics, we engaged human experts to manually label the hallucinated segments of model responses, and examined the consistency of hallucination metrics. Detailed analysis is available in Section \ref{sec:A10} of the appendix. 

\noindent \textbf{Exploring Mitigation Strategies.}
In Section \ref{sec:5.5}, we observe that the OPERA mitigation \cite{huang2024opera}, though designed to counteract the attention sink behaviors, fails to defend against the adversarial visual inputs. One possible reason is that it only focuses on naturally occurring sink tokens during normal generation processes, and reduces its efficacy in detecting and mitigating the adversarial manipulation of attention. Moreover, our attack achieves consistent success across all tested mitigation methods, underscoring their vulnerabilities in solving natural hallucinations but not considering adversarial inputs adequately. We hope this work inspires future research into more robust defensive strategies for MLLMs, such as those based on adversarial purification with diffusion models \cite{lee2023robust, kang2024diffattack}, the defensive system prompt \cite{wang2024adashield} against hallucination, and new training paradigms to overcome the challenges in instruction-tuning stages.

\noindent \textbf{Adversarial Textual Inputs.}
Our hallucination attack crafts adversarial visual inputs to achieve high effectiveness and transferability in MLLMs. Considering their multi-modal nature, exploring the perturbation of textual inputs is also valuable. Existing methods for optimizing adversarial texts typically involve gradient-based searches on predefined target responses and modifying discrete tokens to improve attack success \cite{zou2023universal}. A recent study of jailbreaking attacks against MLLMs suggests decoding adversarial visual inputs within the discrete textual domain \cite{niu2024jailbreaking} to narrow the sampling space. However, the adversarial texts created through these methods generally lack meaningful semantics and are easily detectable. Our approach, on the other hand, optimizes adversarial perturbations by manipulating hidden states and attention mechanisms, which may overcome the current challenges of malicious textual inputs.

\noindent \textbf{Future Works.} In our further research, we plan to extend the adversarial efficacy of the proposed attack to a broader range of multi-modal tasks, including visual reasoning, grounding, and multi-turn visual dialogues. Given the rapid adoption of MLLMs in commercial applications, we will also expand our evaluation to more commercial APIs and downstream modules, as evidence of their multi-modal capabilities. Additionally, investigating mitigation strategies through the lens of attention mechanisms and developing defensive approaches are promising directions for future work.
\section{Conclusion}
This work approaches the prevalent hallucination problem in emerging MLLMs. Through a detailed analysis of the instruction-tuning phase of training, we reveal a critical link between the attention sink phenomenon and hallucinated responses, shedding light on the mechanisms behind erroneous outputs. We propose a novel hallucination attack that induces attention sink behaviors, overcoming the limitations of previous adversarial methods that rely on predefined patterns. Our attack exhibits high transferability, effectively bypassing extensive mitigation strategies and the latest closed-source commercial APIs. We aim to contribute to the safe and reliable development of MLLMs by highlighting current vulnerabilities and inspiring future mitigation strategies.
\section*{Acknowledgement}
We sincerely appreciate the valuable comments from the shepherd and reviewers that improve the paper's quality. 
This work was supported in part by the National Natural Science Foundation of China (62472096, 62172104, 62172105, 62102093, 62102091, 62302101, 62202106).
Min Yang is a faculty of the Shanghai Institute of Intelligent Electronics \& Systems and Engineering Research Center of Cyber Security Auditing and Monitoring, Ministry of Education, China.

\section*{Ethics Considerations}
Our work investigates the hallucination problems in current MLLMs and highlights the potential limitations of existing mitigation strategies. To enhance the faithfulness and helpfulness of MLLMs, we have shared our findings and examples with the providers of commercial MLLM APIs analyzed in this study (e.g., OpenAI, Google) via email.

In all the experiments, the model responses only include hallucinated content that is inconsistent with visual inputs, with no harmful or malicious responses generated. All tests conducted with commercial APIs adhere to the platform's usage guidelines, without any dissemination of hallucinated content or negative impacts on downstream applications. 

To facilitate further research on MLLM hallucination, we release our attack algorithms under a restrictive open-sourcing format. The access will be granted only upon request and exclusively for research purposes, mitigating misuse risks. Additionally, the adversarial visual inputs generated during this study will remain confidential to prevent potential harm.

\section*{Open Science}
We are committed to the principles of open science and have made our source code available upon request for research purposes. Due to the limitations of restrictive open-source licensing, our code is not hosted on a platform with persistent access. Researchers are encouraged to contact us for access to the artifacts via \url{https://huggingface.co/RachelHGF/Mirage-in-the-Eyes}.

\begin{spacing}{0.9}
\bibliographystyle{plain}
\bibliography{main}
\end{spacing} 

\appendix
\appendix

\section{Decoding Strategies of MLLMs}
\label{sec:A1}
Decoding strategies play a vital role in the quality and relevance of MLLM responses, which guide the prediction of the next token based on the distribution of the current generated sequence. Several methods, such as greedy search, beam search, and nucleus sampling, have been proposed to improve text generation quality.

Greedy search \cite{chickering2002optimal}, characterized by its simplicity and computational efficiency, selects the token with the highest probability at each step. However, this approach can lead to sub-optimal results, with the generated text lacking diversity and sometimes getting stuck in repetitive or overly deterministic outputs.

Beam search \cite{holtzman2019curious, xie2024self} is an advanced decoding strategy that attempts to balance response quality and computational cost. Maintaining a fixed number of candidate sequences, known as the beam width, beam search allows MLLMs to explore multiple potential paths during decoding, thereby increasing the likelihood of identifying an optimal sequence. The beam width is defined as a critical hyperparameter, where a larger width expands the search space, but at the cost of increased computational resources.

Nucleus sampling \cite{holtzman2019curious}, or Top-p sampling, introduces a controlled element of randomness to the text generation process to enhance the diversity of output. This method selects the next token from a dynamically determined subset of tokens, where the cumulative probability surpasses a predefined threshold. By adjusting the value of p, nucleus sampling effectively manages the trade-off between randomness and determinism, offering more varied and coherent responses.

\section{Details of Implementation}
\label{sec:A2}

During the hallucination attack, we set the total steps $S_{\rm max}$ of adversarial optimization as 30, with a learning rate $\gamma$ of 5 to update the adversarial perturbations. The attack budget $\epsilon$ is varied across $2/255$, $5/255$, and $8/255$ to modulate the perturbation magnitude. Following an ablation study, we set the hyper-parameter $\alpha=1$ to optimize attack effectiveness. When retrieving hidden states from intermediate levels, we select the second-to-last layer for InstructBLIP and LLaVA-1.5, and the third-to-last layer for MiniGPT-4 and Shikra. During model response generation with beam search decoding, we configured the beam width $N_{\rm beam}$ to 3.
\section{Prompt for GPT-4 Assisted Evaluation}

\begin{figure*}[htbp]
    \centering
\includegraphics[width=\textwidth]{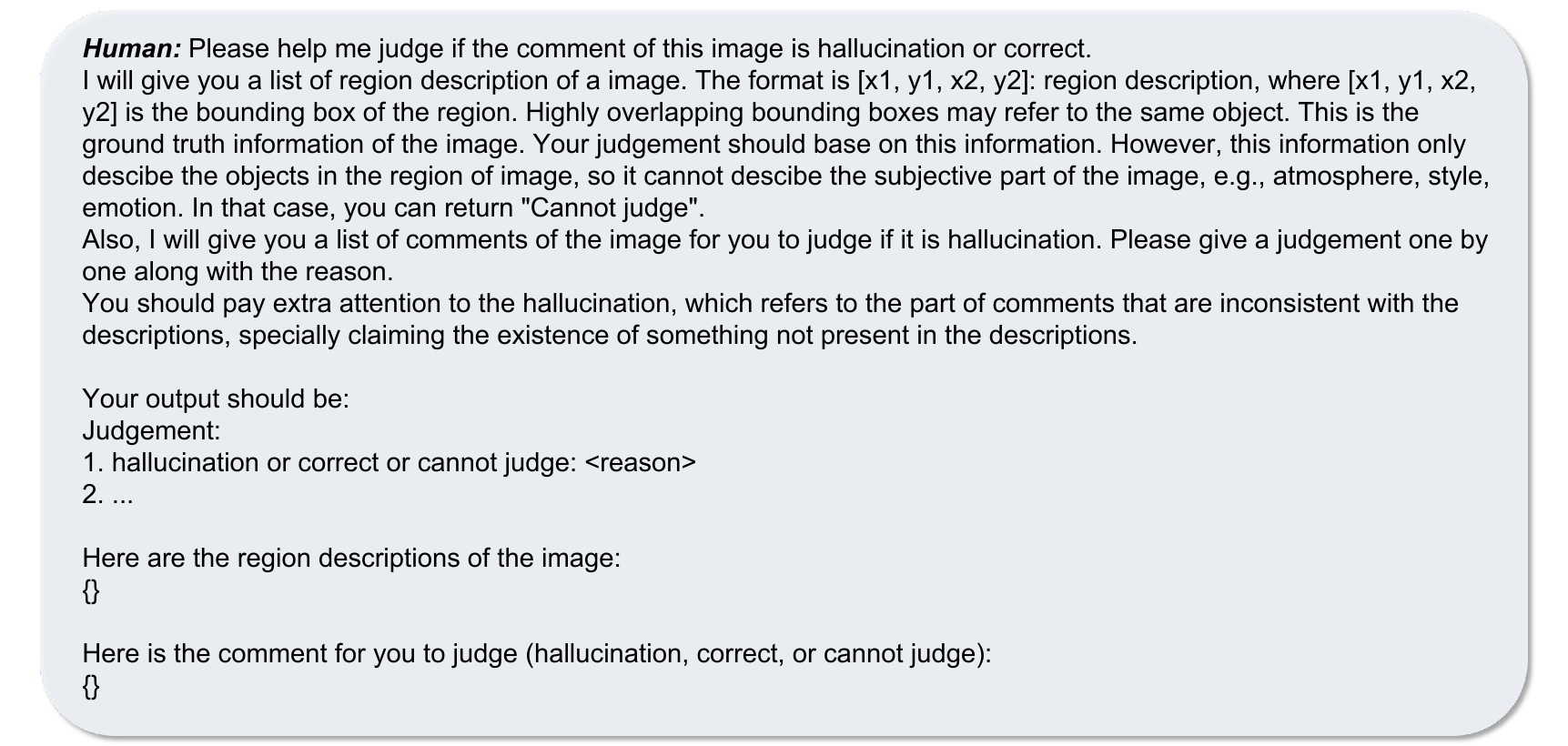}
\caption{The prompt for GPT-4 assisted hallucination evaluation. The bounding boxes coordinates, detailed descriptions, and human-annotated factual knowledge are provided as context information.}
\label{fig:hal_prompt}
\end{figure*}

To assess the hallucination of objects, attributes, and relationships in open-ended model responses, we adopt the GPT-4 assisted evaluation on HalluBench \cite{zhao2023beyond}. The prompt used for hallucination evaluation is displayed in Fig. \ref{fig:hal_prompt}, where the bounding box coordinates, the region descriptions, and human-annotated factual knowledge are provided as a context. 

When performing the hallucination evaluation, we consider six aspects of metrics, which are listed as follows.
\begin{enumerate}
    \item \textbf{SPI}: The number of sentences per image in MLLM responses. This metric quantifies the level of detail in an MLLM’s response at the sentence level.
    \item \textbf{WPI}: The number of words per image in MLLM responses. This metric assesses the level of detail in an MLLM’s response at the word level.
    \item \textbf{HSPI}: The number of hallucinated sentences per image. This metric indicates the extent of hallucination within an MLLM’s response at the sentence level, counting any sentence that contains fabricated content.
    \item \textbf{HWPI}: The number of hallucinated words per image. This metric measures the extent of hallucination within an MLLM’s response at the word level, accounting for any words associated with hallucinated content.
    \item \textbf{HSR}: The ratio of hallucinated sentences. This metric represents the average proportion of hallucinated sentences in the total number of sentences across various MLLM responses on different images.
    \item \textbf{HWR}: The ratio of hallucinated words. This metric captures the average proportion of hallucinated words in the total word count across different MLLM responses on various images.
\end{enumerate}

\begin{figure*}[htbp]
    \centering
\includegraphics[width=0.98\textwidth]{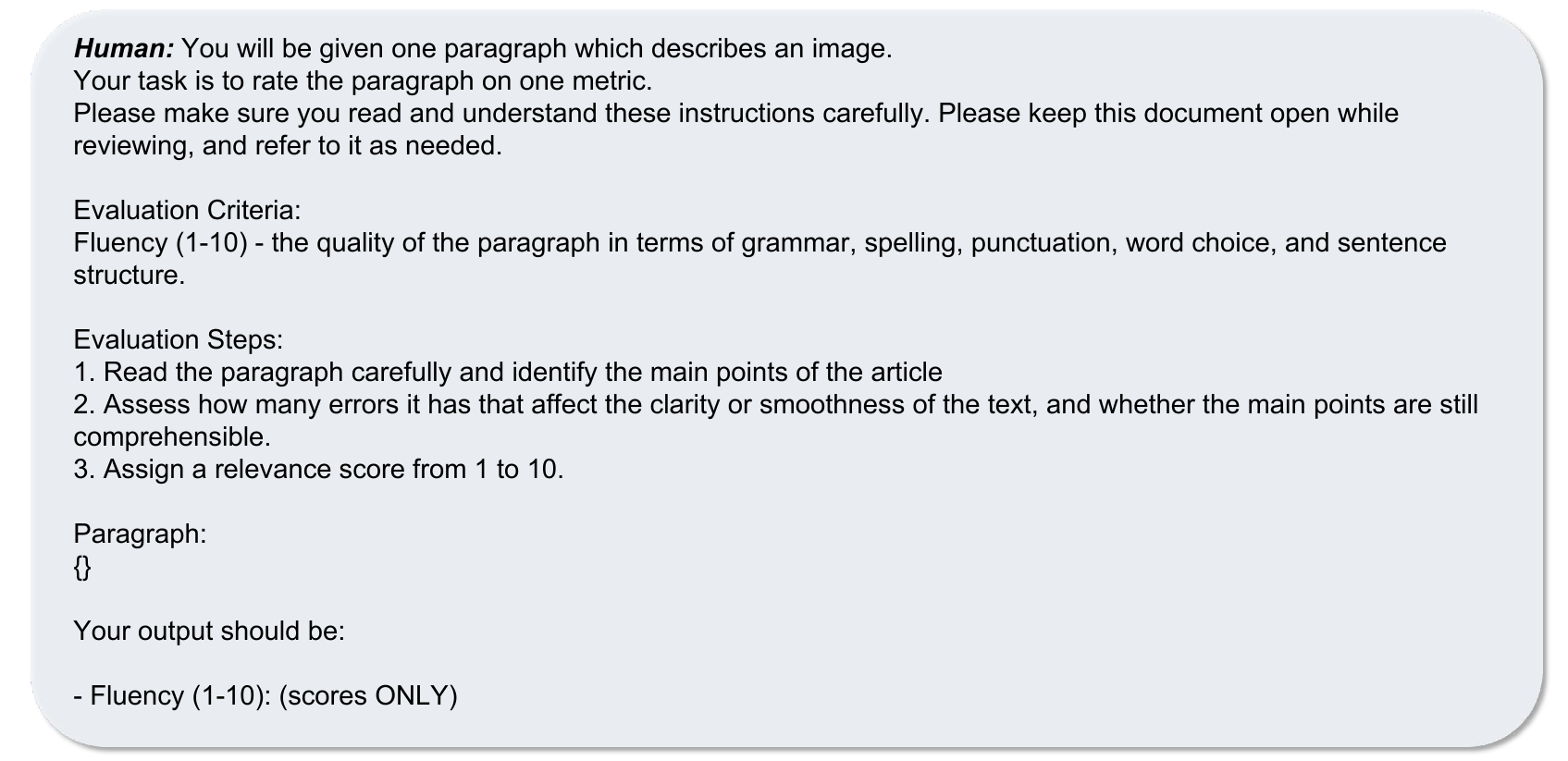}
\caption{The prompt for GPT-4 assisted response quality evaluation. The evaluation criteria, and detailed steps are provided as context information.}
\label{fig:qua_prompt}
\end{figure*}

In addition to the hallucinated content, we also consider the quality of model responses as part of the adversarial goals. We also resort to the GPT-4 model to assess the grammar, fluency, and naturalness of generated responses. The prompt used for response quality evaluation is displayed in Fig. \ref{fig:qua_prompt}.

\section{More Results of Adversarial Effects}

In Section \ref{sec:5.2}, we report GPT-4 assisted hallucination evaluation results on surrogate models using \textit{beam search}. The remaining results for \textit{greedy search}, and \textit{nucleus sampling} decoding are displayed in Tab. \ref{tab:white_g} and Tab. \ref{tab:white_n}.

\begin{table*}[]
\setlength{\belowcaptionskip}{0.15cm}
\caption{Results of GPT-4 assisted hallucination evaluation for the image captioning task on white-box models. All of the MLLM responses are generated with \textit{greedy search} decoding. The six aspects of evaluation are the same as in Tab. \ref{tab:white_b}. A larger HSPI, HWPI, HSR, and HWR indicate a higher level of hallucination in MLLM responses. The best results are marked in bold, and the number in brackets indicates the hallucination improvement compared to the clean image.}
\label{tab:white_g}
\newcommand{\improv}[1]{\footnotesize\textcolor[RGB]{17,74,133}{\textbf{(+#1)}}}
\newcommand{\reduce}[1]{\footnotesize\textcolor[RGB]{81,69,73}{\textbf{(-#1)}}}
\begin{adjustbox}{width=\textwidth}
\begin{tabular}{clllllll}

\toprule[1.5pt]
\makebox[0.15\textwidth][c]{\textbf{Target Model}}   &
\makebox[0.11\textwidth][c]{\textbf{Input}} &
\makebox[0.08\textwidth][l]{\textbf{SPI}} & 
\makebox[0.08\textwidth][l]{\textbf{WPI}} & 
\makebox[0.1\textwidth][l]{\textbf{HSPI}} & 
\makebox[0.1\textwidth][l]{\textbf{HWPI}} & 
\makebox[0.1\textwidth][l]{\textbf{HSR(\%)}} & 
\makebox[0.1\textwidth][l]{\textbf{HWR(\%)}} \\
\midrule

\multirow{4}{*}{\textbf{InstructBLIP}} & clean image   &3.34  &102.89  &2.25  &78.39  &68.05\%  &77.27\%  \\
& $\epsilon$=2/255 &3.32  &103.84  &2.22 \reduce{0.03}  &78.37 \reduce{0.02}  &69.15\% \improv{1.10\%} &\textbf{78.39\%} \improv{1.12\%}  \\
& $\epsilon$=5/255 &3.35  &102.46  &2.30 \improv{0.05}  &78.35 \reduce{0.04}  &70.16\% \improv{2.11\%}  &78.09\% \improv{0.82\%} \\
& $\epsilon$=8/255 &3.38  &98.61  &2.42 \improv{0.17}  &77.03 \reduce{1.36}  &\textbf{71.72\%} \improv{3.67\%} &78.12\% \improv{0.85\%}  \\

\midrule
\multirow{4}{*}{\textbf{LLaVA-1.5}}    & clean image   &5.09  &90.27  &2.27  &42.77  &45.17\% &48.11\%  \\
   & $\epsilon$=2/255 &5.05  &89.91  &2.38 \improv{0.11}  &43.79 \improv{1.02}  &48.63\% \improv{3.46\%}  &50.26\% \improv{2.15\%} \\
   & $\epsilon$=5/255 &5.07  &90.83  &2.38 \improv{0.11}  &45.00 \improv{2.23}  &49.72\% \improv{4.55\%} &52.64\% \improv{4.53\%}  \\
   & $\epsilon$=8/255 &5.07  &90.77  &2.67 \improv{0.40}  &50.23 \improv{7.46}  &\textbf{53.04\%} \improv{7.87\%} &\textbf{55.78\%} \improv{7.67\%} \\

\midrule
\multirow{4}{*}{\textbf{MiniGPT-4}}    & clean image   &5.36  &79.70  &2.92  &44.61  &54.42\%  &56.02\%  \\
   & $\epsilon$=2/255 &5.40  &82.93  &3.19 \improv{0.27}  &50.49 \improv{5.88}  &\textbf{59.37\%} \improv{4.95\%} &\textbf{61.10\%} \improv{5.08\%} \\
   & $\epsilon$=5/255 &5.28  &79.64  &3.02 \improv{0.10}  &47.54 \improv{2.93} &57.65\%  \improv{3.23\%} &60.12\% \improv{4.10\%} \\
   & $\epsilon$=8/255 &5.12  &77.83  &2.92 &46.00 \improv{1.39}  &57.38\% \improv{2.96\%} &59.42\% \improv{3.40\%} \\

\midrule
\multirow{4}{*}{\textbf{Shikra}}   & clean image   &4.99  &91.00  &2.28  &44.09  &45.70\%  &48.58\%  \\
   & $\epsilon$=2/255 &5.02  &90.93  &2.46 \improv{0.18}  &47.03 \improv{2.94}  &51.02\% \improv{5.32\%}  &54.04\% \improv{5.46\%} \\
   & $\epsilon$=5/255 &5.03  &91.95  &2.52 \improv{0.24}  &48.47 \improv{4.38}  &53.83\% \improv{8.13\%} &56.57\% \improv{7.99\%} \\
   & $\epsilon$=8/255 &5.03  &92.53  &2.63 \improv{0.35} &50.68 \improv{6.59} &\textbf{56.02\%} \improv{10.32\%} &\textbf{58.99\%} \improv{10.41\%}  \\
\bottomrule[1.5pt]  
\end{tabular}
\end{adjustbox}
\vspace{-1.5em}
\end{table*}
\begin{table*}[]
\setlength{\belowcaptionskip}{0.15cm}
\caption{Results of GPT-4 assisted hallucination evaluation for the image captioning task on white-box models. All of the MLLM responses are generated with \textit{nucleus sampling} decoding. The six aspects of evaluation are the same as in Tab. \ref{tab:white_b}. A larger HSPI, HWPI, HSR, and HWR indicate a higher level of hallucination in MLLM responses. The best results are marked in bold, and the number in brackets indicates the hallucination 
improvement compared to the clean image.}
\label{tab:white_n}
\newcommand{\improv}[1]{\footnotesize\textcolor[RGB]{17,74,133}{\textbf{(+#1)}}}
\newcommand{\reduce}[1]{\footnotesize\textcolor[RGB]{81,69,73}{\textbf{(-#1)}}}
\begin{adjustbox}{width=\textwidth}
\begin{tabular}{clllllll}

\toprule[1.5pt]
\makebox[0.15\textwidth][c]{\textbf{Target Model}}   &
\makebox[0.11\textwidth][c]{\textbf{Input}} &
\makebox[0.08\textwidth][l]{\textbf{SPI}} & 
\makebox[0.08\textwidth][l]{\textbf{WPI}} & 
\makebox[0.1\textwidth][l]{\textbf{HSPI}} & 
\makebox[0.1\textwidth][l]{\textbf{HWPI}} & 
\makebox[0.1\textwidth][l]{\textbf{HSR(\%)}} & 
\makebox[0.1\textwidth][l]{\textbf{HWR(\%)}} \\
\midrule

\multirow{4}{*}{\textbf{InstructBLIP}} & clean image   &4.98  &90.06  &2.46  &46.46  &49.39\%  &51.61\%  \\
& $\epsilon$=2/255 &5.02  &91.52  &2.55 \improv{0.09} &49.02 \improv{2.56} &51.28\% \improv{1.89\%}  &53.92\% \improv{2.31\%} \\
& $\epsilon$=5/255 &5.03  &91.67  &2.52 \improv{0.06} &48.08 \improv{1.62} &51.73\% \improv{2.34\%} &54.01\% \improv{2.40\%} \\
& $\epsilon$=8/255 &4.97  &90.26  &2.77 \improv{0.31} &52.57 \improv{6.11} &\textbf{56.31\%} \improv{6.92\%} &\textbf{58.92\%} \improv{7.31\%} \\

\midrule
\multirow{4}{*}{\textbf{LLaVA-1.5}}    & clean image   &4.93  &87.32  &2.37  &43.51  &48.03\%  &49.86\%  \\
   & $\epsilon$=2/255 &4.96  &89.49  &2.42 \improv{0.05} &45.33 \improv{1.82}  &49.79\% \improv{1.76\%} &51.80\% \improv{1.94\%} \\
   & $\epsilon$=5/255 &4.88  &88.37  &2.57 \improv{0.20} &48.86 \improv{5.35} &52.91\% \improv{4.88\%} &55.50\% \improv{5.64\%} \\
   & $\epsilon$=8/255 &4.92  &88.58  &2.72 \improv{0.35}  &50.93 \improv{7.42} &\textbf{55.67\%} \improv{7.64\%} &\textbf{57.80\%} \improv{7.94\%} \\

\midrule
\multirow{4}{*}{\textbf{MiniGPT-4}}    & clean image   &4.84  &75.97  &2.56  &41.78  &54.00\%  &56.05\%  \\
   & $\epsilon$=2/255 &4.88  &76.54  &2.80 \improv{0.24} &44.78 \improv{3.0} &58.14\% \improv{4.14\%} &59.35\% \improv{3.30\%} \\
   & $\epsilon$=5/255 &4.80  &75.95  &2.76 \improv{0.20} &44.77 \improv{2.99} &58.11\% \improv{4.11\%} &59.50\% \improv{3.45\%} \\
   & $\epsilon$=8/255 &4.77  &75.20  &2.90 \improv{0.34} &47.07 \improv{5.29} &\textbf{61.27\%} \improv{7.27\%} &\textbf{62.91\%} \improv{6.86\%} \\

\midrule
\multirow{4}{*}{\textbf{Shikra}}   & clean image   &4.77  &86.95  &2.30  &44.12  &48.24\%  &50.71\%  \\
   & $\epsilon$=2/255 &4.85  &87.43  &2.43 \improv{0.13} &46.40 \improv{2.28} &52.30\% \improv{4.06\%} &55.16\% \improv{4.45\%} \\
   & $\epsilon$=5/255 &4.82  &86.10  &2.41 \improv{0.11} &44.79 \improv{0.67} &52.74\% \improv{4.50\%} &54.13\% \improv{3.42\%} \\
   & $\epsilon$=8/255 &4.83  &87.38  &2.61 \improv{0.31} &49.06 \improv{4.94} &\textbf{56.98\%} \improv{8.74\%} &\textbf{59.39\%} \improv{8.68\%} \\
\bottomrule[1.5pt]  
\end{tabular}
\end{adjustbox}
\end{table*}
\section{Results of Human-Evaluated Hallucination Types}
\begin{figure}[!h]
\centering
\includegraphics[width=\columnwidth]{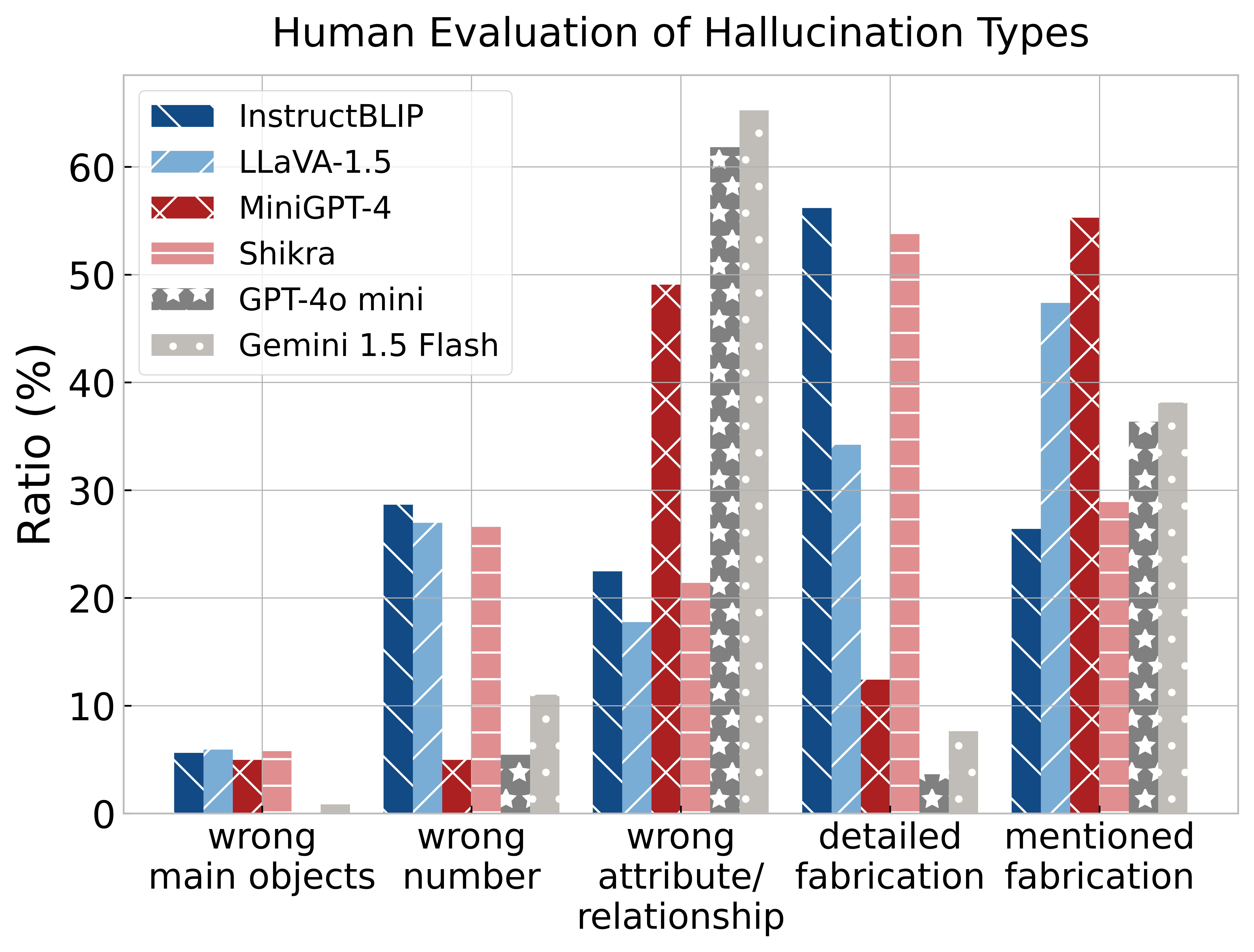}
\vspace{-1.8em}
\caption{Results of the human evaluation on hallucination types. The five types of hallucinations are: identifying the \textit{wrong main objects}, \textit{wrong number} of objects, assigning \textit{wrong attributes or relationships} to objects, generating \textit{detailed fabrication} content, and producing only \textit{mentioned fabrication} content.}
\label{fig:halu_human}
\vspace{-1em}
\end{figure}

To further analyze the adversarial effects of our attack, we engaged human experts to classify the types of hallucinations it induced. The evaluation focused on model responses from the image captioning task across 6 MLLMs in our experiments, with results detailed in Fig. \ref{fig:halu_human}. Open-source MLLMs demonstrate a higher propensity for generating fabricated content when exposed to adversarial visual inputs, whereas commercial APIs more frequently misidentify attributes or relationships between objects.

\section{More Results of Black-box Transferability}

In Section \ref{sec:5.3.1}, we report GPT-4 assisted hallucination evaluation results on black-box models using \textit{beam search}. The remaining results for \textit{greedy search}, and \textit{nucleus sampling} decoding are shown in Tab. \ref{tab:black_g} and Tab. \ref{tab:black_n}.

\begin{figure}[h]
\centering
\includegraphics[width=\columnwidth]{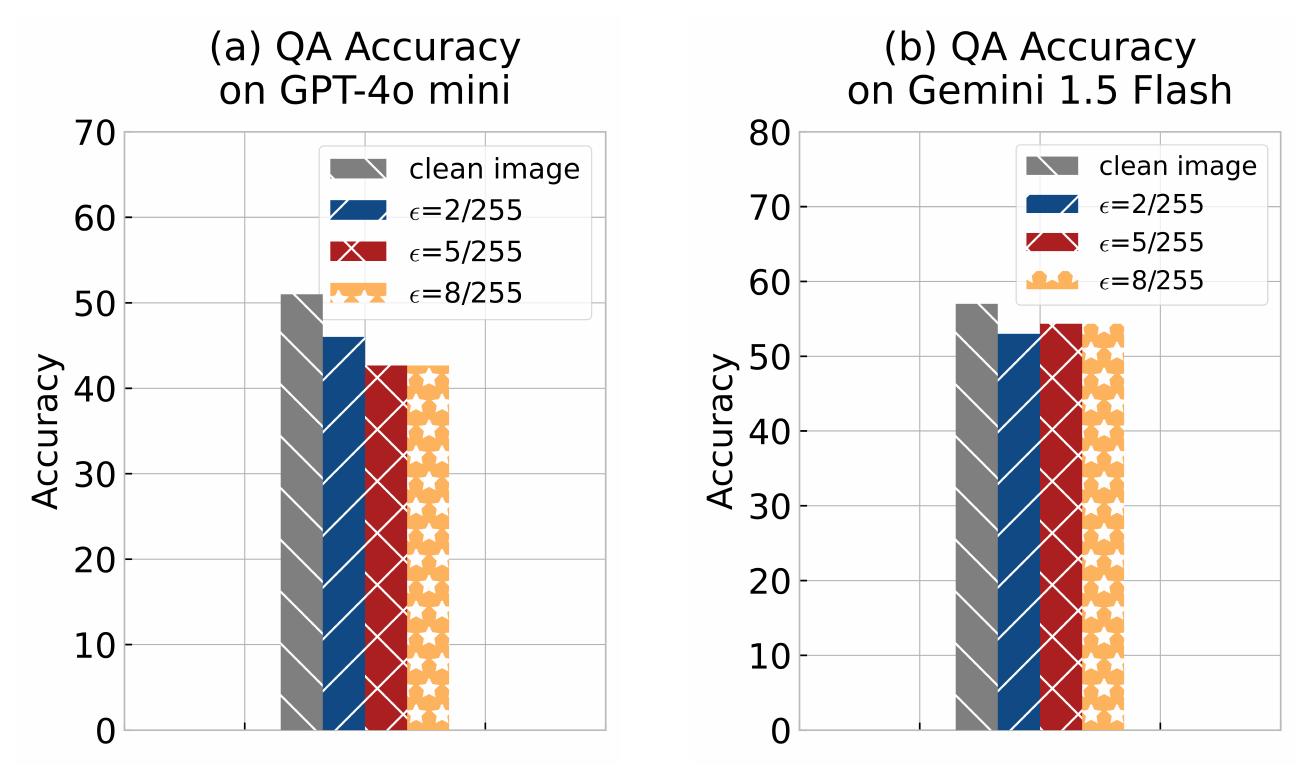}
\vspace{-1.2em}
\caption{Results of QA Accuracy on commercial APIs: (a) GPT-4o mini released by OpenAI and (b) Gemini 1.5 Flash launched by Google. A lower accuracy indicates more hallucinations in MLLM responses.}
\label{fig:api_qa}
\vspace{-1em}
\end{figure}

Additionally, the adversarial QA accuracy results on black-box commercial APIs are illustrated in Fig. \ref{fig:api_qa}. 

\begin{table*}[]
\setlength{\belowcaptionskip}{0.15cm}
\caption{Results of GPT-4 assisted hallucination evaluation for the image captioning task on black-box models. All of the MLLM responses are generated with \textit{greedy search} decoding. The six aspects of evaluation are the same as in Tab. \ref{tab:white_b}. A larger HSPI, HWPI, HSR, and HWR indicate a higher level of hallucination in MLLM responses. The best results are marked in bold, and the number in brackets indicates the hallucination improvement compared to the clean image for each target model.}
\label{tab:black_g}
\newcommand{\improv}[1]{\footnotesize\textcolor[RGB]{17,74,133}{\textbf{(+#1)}}}
\newcommand{\reduce}[1]{\footnotesize\textcolor[RGB]{81,69,73}{\textbf{(-#1)}}}
\begin{adjustbox}{width=\textwidth}
\begin{tabular}{ccllllll}

\toprule[1.5pt]
\makebox[0.15\textwidth][c]{\textbf{Surrogate Model}}   &
\makebox[0.11\textwidth][c]{\textbf{Target Model}} &
\makebox[0.08\textwidth][l]{\textbf{SPI}} & 
\makebox[0.08\textwidth][l]{\textbf{WPI}} & 
\makebox[0.1\textwidth][l]{\textbf{HSPI}} & 
\makebox[0.1\textwidth][l]{\textbf{HWPI}} & 
\makebox[0.1\textwidth][l]{\textbf{HSR(\%)}} & 
\makebox[0.1\textwidth][l]{\textbf{HWR(\%)}} \\
\midrule

\multirow{4}{*}{\textbf{InstructBLIP}} & InstructBLIP   &3.38  &98.61  &2.42 \improv{0.17}  &77.03 \reduce{1.36}  &71.72\% \improv{3.67\%} &78.12\% \improv{0.85\%} \\
& LLaVA-1.5 &5.04  &90.84  &2.43 \improv{0.16} &46.23 \improv{3.46} &48.92\% \improv{3.75\%} &51.62\% \improv{3.51\%} \\
& MiniGPT-4 &5.14  &79.73  &3.18 \improv{0.26} &50.74 \improv{6.13} &\textbf{62.35\%} \improv{7.93\%} &\textbf{64.03\%} \improv{8.01\%} \\
& Shikra &5.11  &93.03  &2.68 \improv{0.40} &51.14 \improv{7.05} &52.69\% \improv{6.99\%} &55.28\% \improv{6.70\%} \\

\midrule
\multirow{4}{*}{\textbf{LLaVA-1.5}}    & LLaVA-1.5   &5.07  &90.77  &2.67 \improv{0.40}  &50.23 \improv{7.46}  &53.04\% \improv{7.87\%} &55.78\% \improv{7.67\%} \\
   & InstructBLIP &3.39  &102.81  &2.41 \improv{0.16} &81.18 \improv{2.79} &71.76\% \improv{3.71\%} &79.86\% \improv{2.59\%} \\
   & MiniGPT-4 &5.37  &80.98  &2.76 \reduce{0.16} &43.51 \reduce{1.10} &56.18\% \improv{1.76\%} &58.17\% \improv{2.15\%} \\
   & Shikra &5.07  &91.66  &2.81 \improv{0.53} &52.88 \improv{8.79} &\textbf{56.38\%} \improv{10.68\%} &\textbf{58.74\%} \improv{10.16\%} \\

\midrule
\multirow{4}{*}{\textbf{MiniGPT-4}}    & MiniGPT-4   &5.40  &82.93  &3.19 \improv{0.27}  &50.49 \improv{5.88}  &59.37\% \improv{4.95\%} &61.10\% \improv{5.08\%} \\
   & InstructBLIP &3.25  &101.19  &2.33 \reduce{0.13} &81.46 \improv{3.07} &72.24\% \improv{4.19\%} &81.21\% \improv{3.94\%} \\
   & LLaVA-1.5 &5.05  &91.03  &2.40 \improv{0.13} &45.86 \improv{3.09} &48.03\% \improv{2.86\%} &51.08\% \improv{2.97\%} \\
   & Shikra &4.98  &90.25  &2.52 \improv{0.24} &47.88 \improv{3.79} &\textbf{50.58\%} \improv{4.88\%} &\textbf{53.10\%} \improv{4.52\%} \\

\midrule
\multirow{4}{*}{\textbf{Shikra}}   
& Shikra   &5.03  &92.53  &2.63 \improv{0.35} &50.68 \improv{6.59} &56.02\% \improv{10.32\%} &58.99\% \improv{10.41\%}  \\
   & InstructBLIP &3.38  &103.03  &2.29 \improv{0.04} &78.95 \improv{0.56} &68.64\% \improv{0.59\%} &77.94\% \improv{0.67\%} \\
   & LLaVA-1.5 &5.07  &90.29  &2.62 \improv{0.35} &48.90 \improv{6.13} &\textbf{52.64\%} \improv{7.47\%} &\textbf{55.19\%} \improv{7.08\%} \\
   & MiniGPT-4 &5.17  &80.17  &2.83 \reduce{0.09} &44.99 \improv{0.38} &56.34\% \improv{1.92\%} &57.95\% \improv{1.93\%} \\
\bottomrule[1.5pt]  
\end{tabular}
\end{adjustbox}
\end{table*}
\begin{table*}[]
\setlength{\belowcaptionskip}{0.15cm}
\caption{Results of GPT-4 assisted hallucination evaluation for the image captioning task on black-box models. All of the MLLM responses are generated with \textit{nucleus sampling} decoding. The six aspects of evaluation are the same as in Tab. \ref{tab:white_b}. A larger HSPI, HWPI, HSR, and HWR indicate a higher level of hallucination in MLLM responses. The best results are marked in bold, and the number in brackets indicates the hallucination improvement compared to the clean image for each target model.}
\label{tab:black_n}
\newcommand{\improv}[1]{\footnotesize\textcolor[RGB]{17,74,133}{\textbf{(+#1)}}}
\newcommand{\reduce}[1]{\footnotesize\textcolor[RGB]{81,69,73}{\textbf{(-#1)}}}
\begin{adjustbox}{width=\textwidth}
\begin{tabular}{ccllllll}

\toprule[1.5pt]
\makebox[0.15\textwidth][c]{\textbf{Surrogate Model}}   &
\makebox[0.11\textwidth][c]{\textbf{Target Model}} &
\makebox[0.08\textwidth][l]{\textbf{SPI}} & 
\makebox[0.08\textwidth][l]{\textbf{WPI}} & 
\makebox[0.1\textwidth][l]{\textbf{HSPI}} & 
\makebox[0.1\textwidth][l]{\textbf{HWPI}} & 
\makebox[0.1\textwidth][l]{\textbf{HSR(\%)}} & 
\makebox[0.1\textwidth][l]{\textbf{HWR(\%)}} \\
\midrule

\multirow{4}{*}{\textbf{InstructBLIP}} & InstructBLIP   &4.97  &90.26  &2.77 \improv{0.31} &52.57 \improv{6.11} &56.31\% \improv{6.92\%} &58.92\% \improv{7.31\%} \\
& LLaVA-1.5 &4.87  &87.15  &2.38 \improv{0.01} &44.93 \improv{1.42} &49.54\% \improv{1.51\%} &52.26\% \improv{2.40\%} \\
& MiniGPT-4 &4.66  &73.85  &2.64 \improv{0.08} &43.17 \improv{1.39} &56.87\% \improv{2.87\%} &58.80\% \improv{2.75\%} \\
& Shikra &4.99  &91.17  &2.71 \improv{0.41} &51.66 \improv{7.54} &\textbf{54.39\%} \improv{6.15\%} &\textbf{56.75\%} \improv{6.04\%} \\

\midrule
\multirow{4}{*}{\textbf{LLaVA-1.5}}    & LLaVA-1.5   &4.92  &88.58  &2.72 \improv{0.35}  &50.93 \improv{7.42} &55.67\% \improv{7.64\%} &57.80\% \improv{7.94\%} \\
   & InstructBLIP &4.96 &90.24  &2.41 \reduce{0.05} &45.81 \reduce{0.65} &51.01\% \improv{1.62\%} &53.68\% \improv{2.07\%} \\
   & MiniGPT-4 &4.47  &73.84  &2.53 \reduce{0.03} &43.68 \improv{1.90} &60.05\% \improv{6.05\%} &62.70\% \improv{6.65\%}\\
   & Shikra &4.84  &87.78  &2.78 \improv{0.48} &52.87 \improv{8.75} &\textbf{57.82\%} \improv{9.58\%} &\textbf{60.66\%} \improv{9.95\%} \\

\midrule
\multirow{4}{*}{\textbf{MiniGPT-4}}    & MiniGPT-4   &4.77  &75.20  &2.90 \improv{0.34} &47.07 \improv{5.29} &61.27\% \improv{7.27\%} &62.91\% \improv{6.86\%} \\
   & InstructBLIP &4.94  &90.88  &2.66 \improv{0.20} &51.33 \improv{4.87} &\textbf{54.08\%} \improv{4.69\%} &\textbf{56.83\%} \improv{5.22\%} \\
   & LLaVA-1.5 &4.89  &87.68  &2.41 \improv{0.04} &46.09 \improv{2.58} &49.38\% \improv{1.35\%} &52.64\% \improv{2.78\%} \\
   & Shikra &4.88  &88.38  &2.42 \improv{0.12} &46.44 \improv{2.32} &49.82\% \improv{1.58\%} &52.80\% \improv{2.09\%} \\

\midrule
\multirow{4}{*}{\textbf{Shikra}}   
& Shikra   &4.83  &87.38  &2.61 \improv{0.31} &49.06 \improv{4.94} &56.98\% \improv{8.74\%} &59.39\% \improv{8.68\%} \\
   & InstructBLIP &4.98  &90.55  &2.51 \improv{0.05} &47.89 \improv{3.43} &52.45\% \improv{3.06\%} &55.31\% \improv{3.70\%} \\
   & LLaVA-1.5 &5.00  &90.84  &2.68 \improv{0.31} &51.08 \improv{7.57} &\textbf{56.63\%} \improv{8.60\%} &\textbf{59.35\%} \improv{9.49\%} \\
   & MiniGPT-4 &4.62  &72.59  &2.45 \reduce{0.11} &39.46 \reduce{2.32} &55.62\% \improv{1.62\%} &57.11\% \improv{1.06\%} \\
\bottomrule[1.5pt]  
\end{tabular}
\end{adjustbox}
\end{table*}
\section{Results of Perplexity-based Model Response Quality}
\begin{figure}[h]
\centering
\includegraphics[width=\columnwidth]{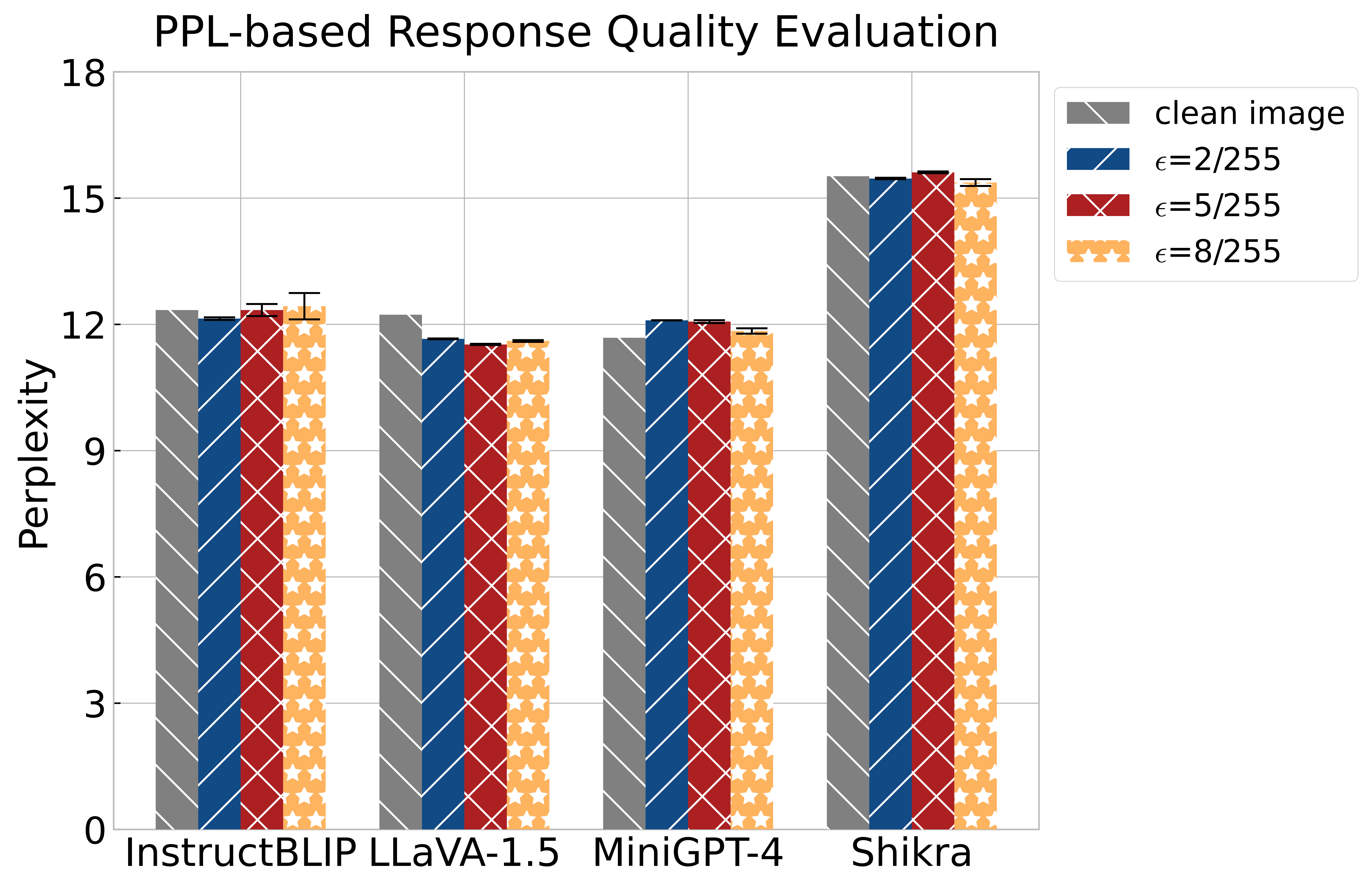}
\caption{Results of Perplexity-based response quality evaluation, covering both white-box and black-box attack scenarios. A lower perplexity reflects better MLLM response quality.}
\label{fig:ppl}
\vspace{-1em}
\end{figure}

In Section \ref{sec:5.4}, we present the results of the GPT-4-assisted evaluation of model response quality. The Perplexity-based quality evaluation results are presented in Fig. \ref{fig:ppl}, calculated with the pre-trained GPT-2 model.

\section{Results of Adaptive Mitigation}
\begin{table}[h]
\setlength{\belowcaptionskip}{0.15cm}
\caption{Results of the length (\textbf{WPI}), hallucination (\textbf{HWR}), and response quality under the early-stopping mitigation strategy.  clean$^{\ast}$ refers to the results of clean images on MLLM without any mitigation strategy applied, while clean$^{\circ}$ indicates the results on MLLM with adaptive mitigation. The best adversarial results are marked in bold.}
\label{tab:adaptive}
\newcommand{\improv}[1]{\footnotesize\textcolor[RGB]{17,74,133}{\textbf{(+#1)}}}
\newcommand{\reduce}[1]{\footnotesize\textcolor[RGB]{81,69,73}{\textbf{(-#1)}}}

\centering
\begin{adjustbox}{width=\columnwidth}
\begin{tabular}{ccllll}
\toprule[1.5pt]
\makebox[0.05\columnwidth][c]{\textbf{Target Model}} & \makebox[0.03\columnwidth][c]{\textbf{Input}}  & \makebox[0.08\columnwidth][l]{\textbf{WPI}} & \makebox[0.1\columnwidth][l]{\textbf{HWR(\%)}} & \makebox[0.12\columnwidth][l]{\textbf{Quality}} \\ 
\midrule
\multirow{5}{*}{\textbf{InstructBLIP}} & clean$^{\ast}$ & 102.89 & 77.27\% & 8.58 \\ \cline{2-5}
 & clean$^{\circ}$ & 46.51 & 52.56\% & 5.55 \\
 & $\epsilon$=2/255 & 48.57 & 63.08\% \improv{10.52\%} & 5.45 \\
 & $\epsilon$=5/255 & 45.31 & 65.24\% \improv{12.68\%}& 5.57 \\
 & $\epsilon$=8/255 & 46.31 & \textbf{69.99}\% \improv{17.43\%}& 5.28 \\ \hline
\multirow{5}{*}{\textbf{MiniGPT-4}} & clean$^{\ast}$ & 79.70 & 56.02\% & 8.81 \\ \cline{2-5}
 & clean$^{\circ}$ & 51.05 & 53.67\% & 6.39 \\
 & $\epsilon$=2/255 & 50.81 & 58.93\% \improv{5.26\%} & 6.36 \\
 & $\epsilon$=5/255 & 52.47 & 56.64\% \improv{2.97\%} & 6.49 \\
 & $\epsilon$=8/255 & 51.88 & \textbf{59.87}\% \improv{6.20\%} & 5.69 \\  
\bottomrule[1.5pt] 
\end{tabular}
\end{adjustbox}
\vspace{-0.5em}
\end{table}

Considering the adaptive mitigation strategy of detecting and early-stopping before sink tokens, we present the length, quality, and hallucination metrics of MLLMs responses under the mitigation strategy in Tab. \ref{tab:adaptive}.
\section{More Results of Baseline Comparison}

In Section \ref{sec:5.6}, we report GPT-4 assisted hallucination evaluation results on baseline methods using \textit{beam search}. The remaining results for \textit{greedy search}, and \textit{nucleus sampling} decoding are shown in Tab. \ref{tab:baseline_g} and Tab. \ref{tab:baseline_n}.

\begin{table*}[h]
\setlength{\belowcaptionskip}{0.15cm}
\caption{Results of GPT-4 assisted hallucination attack of the baseline method with \textit{greedy sampling} decoding. $\delta$ denotes of budget of random noises injected into visual inputs. The line of \textit{attack} denotes the best results in the white-box attack scenario.}
\label{tab:baseline_g}
\newcommand{\improv}[1]{\scriptsize\textcolor[RGB]{17,74,133}{\textbf{(+#1)}}}
\newcommand{\reduce}[1]{\scriptsize\textcolor[RGB]{81,69,73}{\textbf{(-#1)}}}

\begin{adjustbox}{width=\textwidth}
\begin{tabular}{cllllllll}
\toprule[1.5pt]
\makebox[0.02\textwidth][c]{} & \multicolumn{2}{l}{\ \ \ \ \ \ \ \ \ \ \textbf{InstructBLIP}} & 
\multicolumn{2}{l}{\ \ \ \ \ \ \ \ \ \ \ \ \ \textbf{LLaVA-1.5}} & \multicolumn{2}{l}{\ \ \ \ \ \ \ \ \ \ \ \ \ \textbf{MiniGPT-4}}        & \multicolumn{2}{l}{\ \ \ \ \ \ \ \ \ \ \ \ \ \ \ \  \textbf{Shikra}}   \\
\midrule
& \makebox[0.02\textwidth][l]{\textbf{HSR(\%)}}    & \makebox[0.02\textwidth][l]{\textbf{HWR(\%)}}    & \makebox[0.02\textwidth][l]{\textbf{HSR(\%)}}    & \makebox[0.02\textwidth][l]{\textbf{HWR(\%)}}    & \makebox[0.02\textwidth][l]{\textbf{HSR(\%)}}    & \makebox[0.02\textwidth][l]{\textbf{HWR(\%)}}    & \makebox[0.02\textwidth][l]{\textbf{HSR(\%)}}    & \makebox[0.02\textwidth][l]{\textbf{HWR(\%)}}  \\
\midrule
clean & 68.05\% & 77.27\% & 45.17\% & 48.11\% & 54.42\% & 56.02\% & 45.70\% & 48.58\%  \\
$\delta$=2/255 &65.84\% \reduce{2.21\%} &75.02\% \reduce{2.26\%} &43.73\% \reduce{1.44\%} &45.92\% \reduce{2.19\%} &52.87\% \reduce{1.55\%} &55.15\% \reduce{0.87\%} &47.87\% \improv{2.17\%} &50.27\% \improv{1.69\%}  \\
$\delta$=5/255 &67.74\% \reduce{0.31\%} &77.17\% \reduce{0.10\%} &45.96\% \improv{0.79\%} &48.56\% \improv{0.45\%}  &56.65\% \improv{2.23\%} &58.81\% \improv{2.79\%} &43.50\% \reduce{2.20\%} &46.12\% \reduce{2.46\%} \\
$\delta$=8/255 &63.01\% \reduce{5.04\%} &73.66\% \reduce{3.61\%} &45.84\% \improv{0.67\%} &48.33\% \improv{0.22\%} &55.31\% \improv{0.89\%} &57.30\% \improv{1.28\%} &47.07\% \improv{1.37\%} &49.85\% \improv{1.27\%} \\
\midrule
attack  &\textbf{71.72\%} \improv{3.67\%} &\textbf{78.39\%} \improv{1.12\%} &\textbf{53.04\%} \improv{7.87\%} &\textbf{55.78\%} \improv{7.67\%} &\textbf{59.37\%} \improv{4.95\%} &\textbf{61.10\%} \improv{5.08\%} &\textbf{56.02\%} \improv{10.32\%} &\textbf{58.99\%} \improv{10.41\%}  \\
\bottomrule[1.5pt]
\end{tabular}
\end{adjustbox}
\end{table*}
\begin{table*}[h]
\setlength{\belowcaptionskip}{0.15cm}
\caption{Results of GPT-4 assisted hallucination attack of the baseline method with \textit{nucleus sampling} decoding. $\delta$ denotes of budget of random noises injected into visual inputs. The line of \textit{attack} denotes the best results in the white-box attack scenario.}
\newcommand{\improv}[1]{\scriptsize\textcolor[RGB]{17,74,133}{\textbf{(+#1)}}}
\newcommand{\reduce}[1]{\scriptsize\textcolor[RGB]{81,69,73}{\textbf{(-#1)}}}
\label{tab:baseline_n}
\begin{adjustbox}{width=\textwidth}
\begin{tabular}{cllllllll}
\toprule[1.5pt]
\makebox[0.02\textwidth][c]{} & \multicolumn{2}{l}{\ \ \ \ \ \ \ \ \ \ \textbf{InstructBLIP}} & 
\multicolumn{2}{l}{\ \ \ \ \ \ \ \ \ \ \ \ \ \textbf{LLaVA-1.5}} & \multicolumn{2}{l}{\ \ \ \ \ \ \ \ \ \ \ \ \ \textbf{MiniGPT-4}}        & \multicolumn{2}{l}{\ \ \ \ \ \ \ \ \ \ \ \ \ \ \ \  \textbf{Shikra}}   \\
\midrule
& \makebox[0.02\textwidth][l]{\textbf{HSR(\%)}}    & \makebox[0.02\textwidth][l]{\textbf{HWR(\%)}}    & \makebox[0.02\textwidth][l]{\textbf{HSR(\%)}}    & \makebox[0.02\textwidth][l]{\textbf{HWR(\%)}}    & \makebox[0.02\textwidth][l]{\textbf{HSR(\%)}}    & \makebox[0.02\textwidth][l]{\textbf{HWR(\%)}}    & \makebox[0.02\textwidth][l]{\textbf{HSR(\%)}}    & \makebox[0.02\textwidth][l]{\textbf{HWR(\%)}}  \\
\midrule
clean & 49.39\% & 51.61\% & 48.03\% & 49.86\% & 54.00\% & 56.05\% & 48.24\% & 50.71\%  \\
$\delta$=2/255 &47.94\% \reduce{1.45\%} &51.44\% \reduce{0.17\%} &48.48\% \improv{0.45\%} &51.19\% \improv{1.33\%} &55.52\% \improv{1.52\%} &56.96\% \improv{0.91\%} &46.32\% \reduce{1.92\%} &48.43\% \reduce{2.28\%}  \\
$\delta$=5/255 &46.92\% \reduce{2.47\%} &49.50\% \reduce{2.11\%} &49.87\% \improv{1.84\%} &53.22\% \improv{3.36\%} &54.44\% \improv{0.44\%} &55.45\% \reduce{0.60\%} &46.52\% \reduce{1.72\%} &49.44\% \reduce{1.27\%} \\
$\delta$=8/255 &46.62\% \reduce{2.77\%} &47.94\% \reduce{3.67\%} &43.29\% \reduce{4.74\%} &46.02\% \reduce{3.84\%} &53.23\% \reduce{0.77\%} &53.85\% \reduce{2.20\%} &46.83\% \reduce{1.41\%} &49.71\% \reduce{1.00\%} \\
\midrule
attack  &\textbf{56.31\%} \improv{6.92\%} &\textbf{58.92\%} \improv{7.31\%} &\textbf{55.67\%} \improv{7.64\%} &\textbf{57.80\%} \improv{7.94\%} &\textbf{61.27\%} \improv{7.27\%} &\textbf{62.91\%} \improv{6.86\%} &\textbf{56.98\%} \improv{8.74\%}  &\textbf{59.39\%} \improv{8.68\%} \\
\bottomrule[1.5pt]
\end{tabular}
\end{adjustbox}
\end{table*}
\section{Results of Alignment Study}
\label{sec:A10}
\begin{table*}[h!]
\setlength{\belowcaptionskip}{0.15cm}
\caption{Comparison of human-evaluated and GPT-4 assisted hallucination metrics on 4 sets of model responses. $(^{\dagger})$ denotes the evaluation results of human experts, while $(^{\circ})$ denotes the evaluation results of GPT-4. The human-evaluated results are averaged between two experts. The number in brackets indicates the hallucination improvement compared to the clean image, with human-evaluated results in \textcolor[RGB]{190,0,47}{\textbf{red color}} and GPT-4-evaluated results in \textcolor[RGB]{17,74,133}{\textbf{indigo color}}. Adversarial results are marked in bold.}
\label{tab:align}
\newcommand{\gptimprov}[1]{\scriptsize\textcolor[RGB]{17,74,133}{\textbf{(+#1)}}}
\newcommand{\improv}[1]{\scriptsize\textcolor[RGB]{190,0,47}{\textbf{(+#1)}}}

\centering
\begin{adjustbox}{width=0.65\textwidth}
\begin{tabular}{ccllll}
\toprule[1.5pt]
   \makebox[0.12\textwidth][c]{} &
   \makebox[0.12\textwidth][c]{\textbf{Input}} & 
   \makebox[0.08\textwidth][l]{\textbf{HSPI}} & 
   \makebox[0.08\textwidth][l]{\textbf{HWPI}} & 
   \makebox[0.1\textwidth][l]{\textbf{HSR(\%)}} & \makebox[0.1\textwidth][l]{\textbf{HWR(\%)}} \\
\midrule
\multirow{4}{*}{\textbf{MiniGPT-4}} & clean image$^{\dagger}$ & 2.53 & 39.84 & 49.32\% & 51.21\% \\
   & clean image$^{\circ}$ & 2.92 & 44.61 & 54.42\% & 56.02\% \\
   & $\epsilon$=8/255$^{\dagger}$ & \textbf{2.79} \improv{0.26} & \textbf{43.04} \improv{3.20} & \textbf{52.10\%} \improv{2.78\%} & \textbf{54.00\%} \improv{2.79\%} \\
   & $\epsilon$=8/255$^{\circ}$ & \textbf{2.92} & \textbf{46.00} \gptimprov{1.39} & \textbf{57.38\%} \gptimprov{2.96\%} & \textbf{59.42\%} \gptimprov{3.40\%} \\
\midrule
\multirow{4}{*}{\textbf{LLaVA-1.5}} & clean image$^{\dagger}$ & 2.16 & 41.31 & 42.44\% & 45.76\% \\
   & clean image$^{\circ}$ & 2.27 & 42.77 & 45.17\% & 48.11\% \\
   & $\epsilon$=8/255$^{\dagger}$ & \textbf{2.41} \improv{0.25} & \textbf{46.62} \improv{5.31} & \textbf{47.58\%} \improv{5.14\%} & \textbf{51.36\%} \improv{5.60\%} \\
   & $\epsilon$=8/255$^{\circ}$ & \textbf{2.67} \gptimprov{0.40} & \textbf{50.23} \gptimprov{7.46} & \textbf{53.04\%} \gptimprov{7.87\%} & \textbf{55.78\%} \gptimprov{7.67\%}  \\
\toprule[1.5pt]
\end{tabular}
\end{adjustbox}
\end{table*}
To assess the alignment between GPT-4-assisted hallucination evaluation and human judgments, we engaged human experts to manually label the hallucinated segments of model responses. We compare the HSPI/HWPI/HSR/HWR metrics derived from GPT-4-based evaluations with those from human assessments. The detailed results are presented in Tab. \ref{tab:align}. Notably, the human-evaluated hallucination metrics are slightly lower than those obtained from GPT-4, likely due to the limited scope of descriptions and the factual information annotated in the HalluBench dataset. Additionally, the results for adversarial visual inputs show comparable improvements, underscoring the reliability and effectiveness of GPT-4-assisted evaluation in this study.
\section{Qualitative Results}

\begin{figure*}[h]
    \centering
\includegraphics[width=\textwidth]{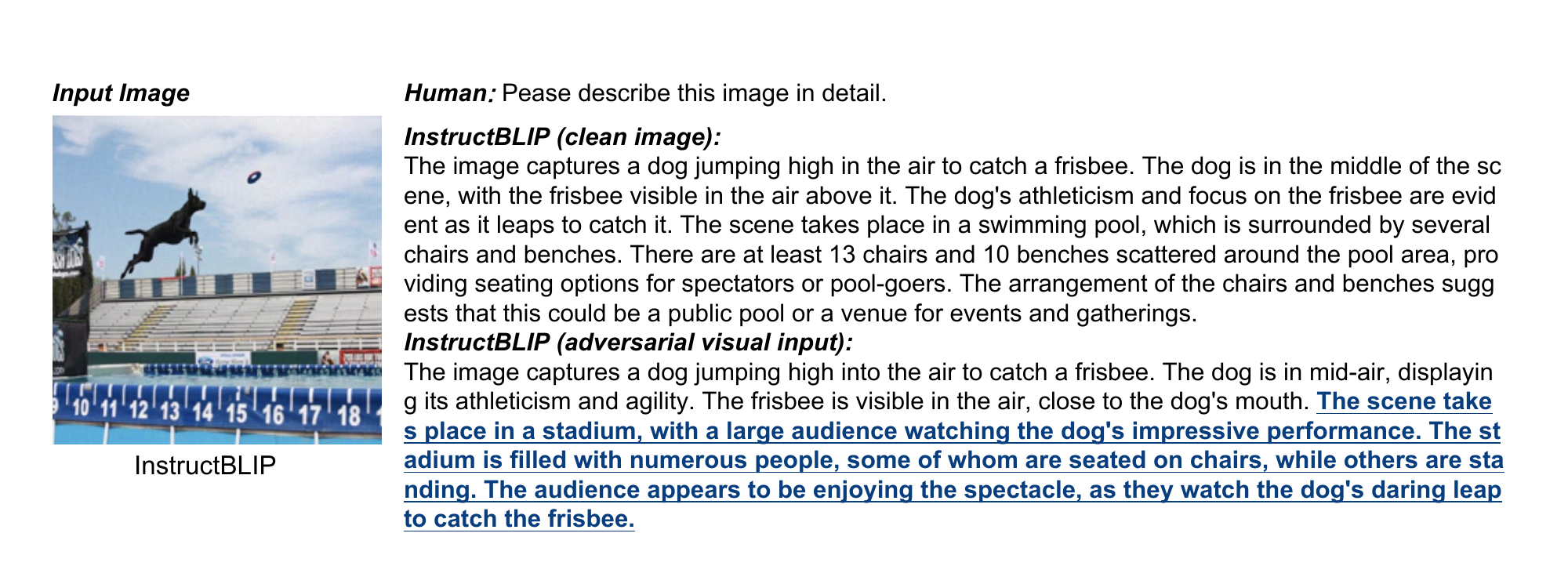}
\vspace{-2em}
\caption{A case of model responses with clean image and adversarial visual input on InstructBLIP.}
\label{fig:case1}
\vspace{-2em}
\end{figure*}

\begin{figure*}[h]
    \centering
\includegraphics[width=\textwidth]{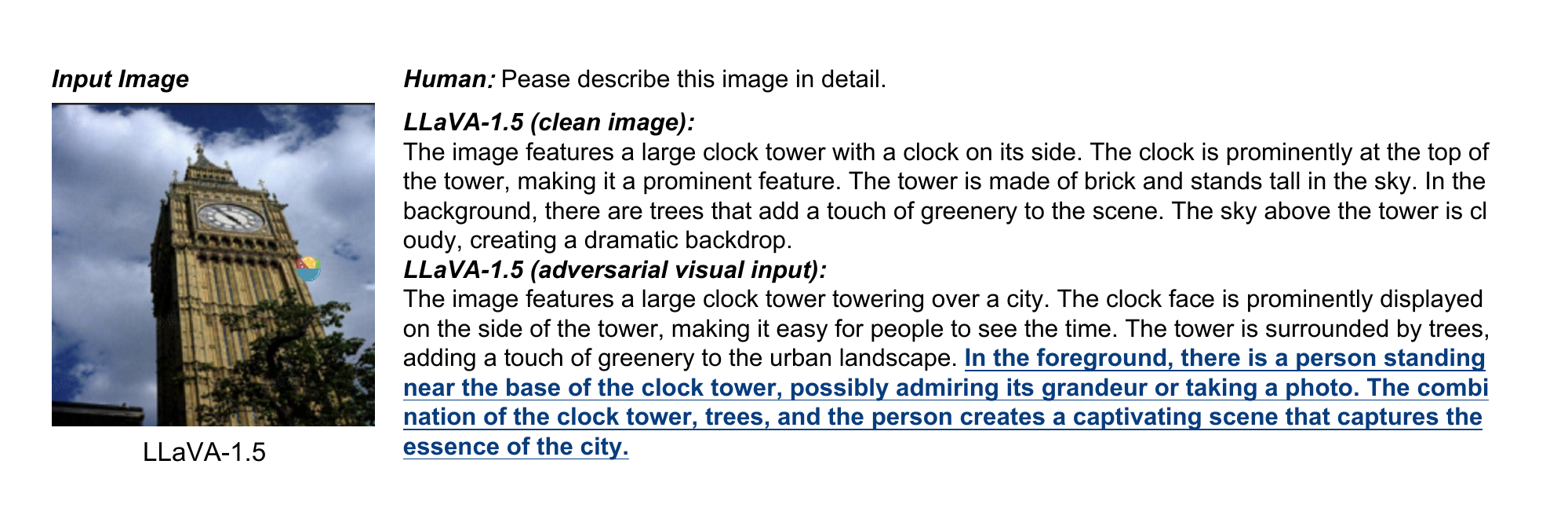}
\vspace{-2em}
\caption{A case of model responses with clean image and adversarial visual input on LLaVA-1.5.}
\label{fig:case2}
\vspace{-2em}
\end{figure*}

\begin{figure*}[!t]
    \centering
\includegraphics[width=\textwidth]{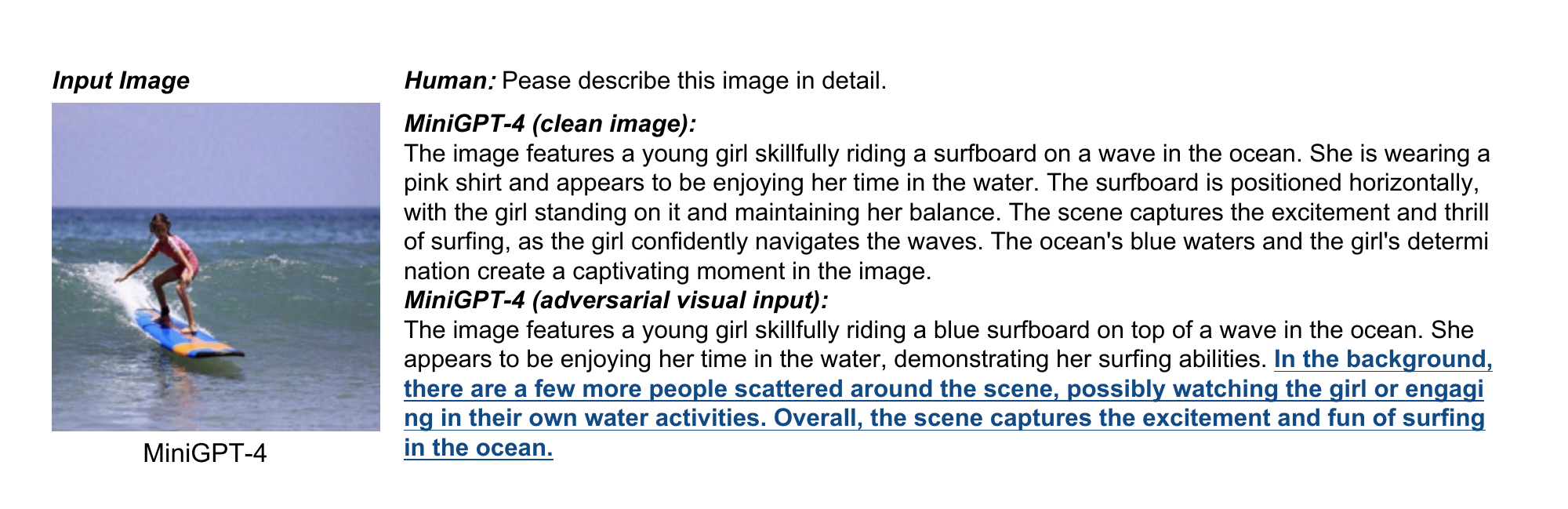}
\vspace{-2em}
\caption{A case of model responses with clean image and adversarial visual input on MiniGPT-4.}
\label{fig:case3}
\vspace{-2em}
\end{figure*}

\begin{figure*}[!t]
    \centering
\includegraphics[width=\textwidth]{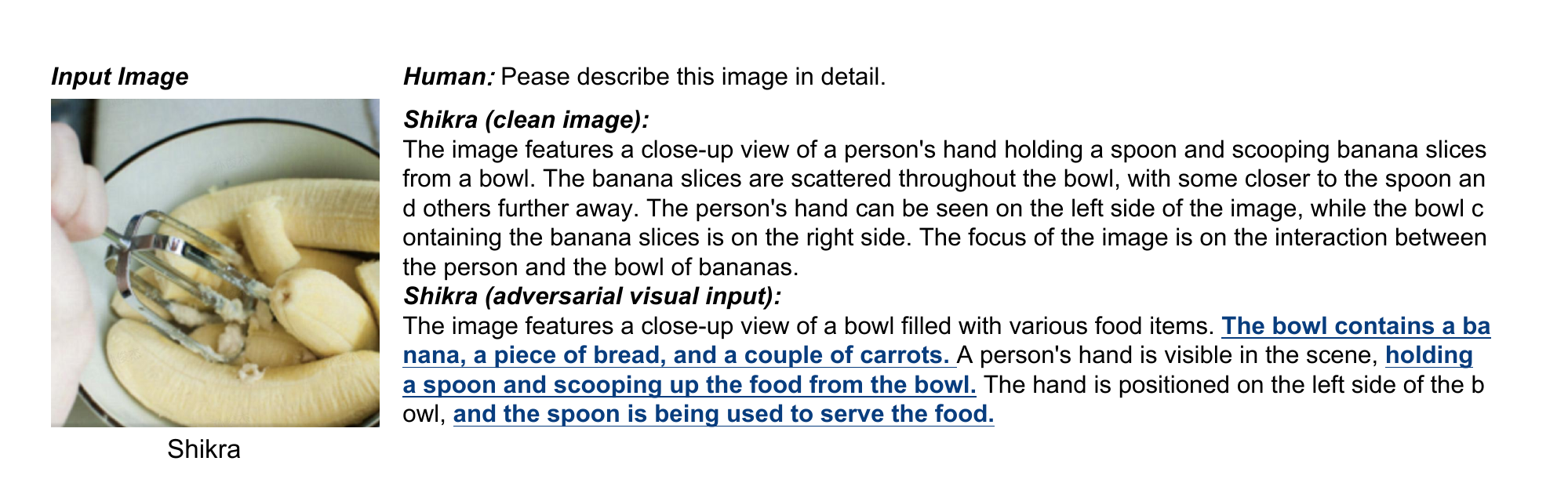}
\vspace{-2em}
\caption{A case of model responses with clean image and adversarial visual input on Shikra.}
\label{fig:case4}
\vspace{-2em}
\end{figure*}

To illustrate the adversarial effects of our proposed attack, we provide qualitative cases comparing the MLLM responses with clean images and adversarial visual inputs in Fig. \ref{fig:case1}, \ref{fig:case2}, \ref{fig:case3}, and \ref{fig:case4}. The examples are chosen from our experiments, with hallucinated content marked bold with \textcolor[RGB]{8,61,126}{\textbf{indigo color}}.

\end{document}